\documentclass[twoside,11pt]{article}

\usepackage{jmlr2e}

\usepackage[active]{srcltx}

\usepackage[utf8]{inputenc}
\usepackage{amsmath}
\usepackage{amssymb}
\usepackage{enumerate}
\usepackage{multirow}
\usepackage{array}
\usepackage{color}
\usepackage{xspace}
\usepackage{booktabs}

\usepackage{url}
\usepackage[safe]{tipa}

\usepackage{slashbox}

\newtheorem{de}{Definition}
\newtheorem{theo}{Theorem}
 
\newtheorem{corrol}[theo]{Corollary}

\newtheorem{rem}{Remark}

\newtheorem{hypothesis}{Assumption}

\newcommand{\N}{\mathbb{N}}
\newcommand{\R}{\mathbb{R}}

\newcommand{\h}{\mathcal{H}}

\newcommand {\x} {\mathcal{X}}
\newcommand {\y} {\mathcal{Y}}
\newcommand {\I} {\mathcal{I}}
\newcommand {\zi} {{Z^i}}
\renewcommand {\l} {\ell}

\usepackage{mdwlist}

\usepackage{tkz-tab}
\usetikzlibrary{decorations.pathreplacing} 

\usepackage[all]{xy}

\usepackage[babel=true]{csquotes}

\newcommand{\eqdef}{\stackrel{\rm\scriptsize kern.\hspace{0.05cm}def}{=}}
\usepackage{algorithm}
\usepackage{algorithmic}

\newcommand{\eg}{\emph{e.g.}\@\xspace}

\definecolor{MyDarkBlue}{rgb}{0.05,0.,0.8}

\jmlrheading{16}{2015}{1-54}{10/11; Revised ??/15}{??/15}
{Hachem Kadri, Emmanuel Duflos, Philippe Preux, St\'ephane Canu, Alain Rakotomamonjy and Julien Audiffren}

\ShortHeadings{Operator-valued Kernels for Learning from Functional Response Data}
{Kadri, Duflos, Preux, Canu, Rakotomamonjy and Audiffren}
\firstpageno{1}

\begin{document}

\title{Operator-valued Kernels for Learning from Functional Response Data}

\author{\name Hachem Kadri \email hachem.kadri@lif.univ-mrs.fr \\
       \addr Aix-Marseille Universit\'e, LIF (UMR CNRS 7279)\\
       F-13288 Marseille Cedex 9, France
	 \AND
	\name Emmanuel Duflos \email emmanuel.duflos@ec-lille.fr \\
       \addr Ecole Centrale de Lille, CRIStAL (UMR CNRS 9189) \\
       59650 Villeneuve d'Ascq, France
 \AND
	\name Philippe Preux \email philippe.preux@univ-lille3.fr \\
       \addr Université de Lille, CRIStAL (UMR CNRS 9189) \\
       59650 Villeneuve d'Ascq, France
 \AND
	\name St\'ephane Canu \email scanu@insa-rouen.fr \\
       \addr INSA de Rouen, LITIS (EA 4108)\\
       76801, St Etienne du Rouvray, France
\AND
	\name Alain Rakotomamonjy \email alain.rakotomamonjy@insa-rouen.fr \\
       \addr Universit\'e de Rouen, LITIS (EA 4108)\\
       76801, St Etienne du Rouvray, France
\AND
	\name Julien Audiffren \email julien.audiffren@cmla.ens-cachan.fr \\
       \addr ENS Cachan, CMLA (UMR CNRS 8536)\\
       94235 Cachan Cedex, France
}

\editor{John Shawe-Taylor}

\maketitle

\begin{abstract}%
In this paper\footnote{This is a combined and expanded version of previous conference
papers~\citep{Kadri-2010,Kadri-2011c}.} we consider the problems of supervised classification and regression in the case where attributes and labels are functions: a data is represented by a set of functions, and the label is also a function. We focus on the use of reproducing kernel Hilbert space theory to
learn from such functional data. Basic concepts and properties of kernel-based learning
are extended to include the estimation of function-valued functions. In this setting, the
representer theorem is restated, a set of rigorously defined infinite-dimensional operator-valued kernels that
can be valuably applied when the data are functions is described, and a learning algorithm for nonlinear functional data analysis is introduced.
The methodology is illustrated through speech and audio signal processing experiments. 
\end{abstract}


\section{Introduction}
\label{sec:intro}

  In this paper, we consider the supervised learning problem in a
  functional setting: each attribute of a data is a function, and the
  label of each data is also a function. For the sake of simplicity,
  one may imagine real functions, though the work presented here is
  much more general; one may also think about those functions as being
  defined over time, or space, though again, our work is not tied to
  such assumptions and is much more general.  To this end, we extend
  the traditional scalar-valued attribute setting to a function-valued
  attribute setting.

  This shift from scalars to functions is required by the simple fact
  that in many applications, attributes are functions: functions may
  be one dimensional such as economic curves~(variation of the price
  of ``actions''), load curve of a server, a sound, etc., or two or
  higher dimensional~(hyperspectral images, etc.). Due to the nature of
  signal acquisition, one may consider that in the end, a signal is
  always acquired in a discrete fashion, thus providing a real
  vector. However, with the resolution getting finer and finer in many
  sensors, the amount of discrete data is getting huge, and one may
  reasonably wonder whether a functional point of view may not be
  better than a vector point of view. Now, if we keep aside the
  application point of view, the study of functional attributes may
  simply come as an intellectual question which is interesting for its
  own sake.

  From a mathematical point of view, the shift from scalar attributes
  to function attributes will come as a generalization from
  scalar-valued functions to function-valued functions, a.k.a.``operators''. Reproducing Kernel Hilbert Spaces (RKHS) has become a
  widespread tool to deal with the problem of learning a function
  mapping the set $\mathbb{R}^p$ to the set of real numbers
  $\mathbb{R}$. Here, we have to deal with RKHS of operators, that are
  functions that map a function, belonging to a certain space of
  functions, to a function belonging to an other space of
  functions. This shift in terminology is accompanied with a dramatic
  shift in concepts, and technical difficulties that have to be
  properly handled.

This \textit{functional regression problem}, or \textit{functional supervised
 learning}, is a challenging research problem, from statistics to machine 
learning. 
Most previous work has focused on the discrete case: the multiple-response 
(finite and discrete) function estimation problem.
In the machine
learning literature, this problem is better known under the name of
vector-valued function learning \citep{Micchelli-2005a}, while in the
field of statistics, researchers prefer to use the term multiple
output regression \citep{Breiman-1997}. One possible solution is to
approach the problem from a univariate point of view, that is,
assuming only a single response variable output from the same set of
explanatory variables. However it would be more efficient to take
advantage of correlation between the response variables by considering
all responses simultaneously. For further discussion of this point, we
refer the reader to \citet{Hastie-2001} and references therein. More
recently, relevant works in this context concern regularized
regression with a sequence of ordered response variables.
Many variable selection and shrinkage methods
for single response regression are extended to the multiple response
data case and several algorithms following the corresponding
solution paths are proposed \citep{Turlach-2005, Simila-2007, Hesterberg-2008}.

Learning from multiple responses is closely related to the problem of
multi-task learning where the goal is to improve generalization
performance by learning multiple tasks simultaneously.  There is a
large literature on this subject, in particular \citet{Evgeniou-2004,
  Jebara-2004, Ando-2005, Maurer-2006, Ben-david-2008, Argyriou-2008}
and references therein. One paper that has come to our attention is
that of \citet{Evgeniou-2005} who showed how Hilbert spaces of
vector-valued functions \citep{Micchelli-2005a} and matrix-valued
reproducing kernels \citep{Micchelli-2005b, Reisert-2007} can be used
as a theoretical framework to develop nonlinear multi-task learning
methods.

A primary motivation for this paper is to build on these previous
studies and provide a similar framework for addressing the general
case where the output space is infinite dimensional.  In this setting,
the output space is a space of functions and elements of this space
are called functional response data. Functional responses are
frequently encountered in the analysis of time-varying data when
repeated measurements of a continuous response variable are taken over
a small period of time~\citep{Faraway-1997, Yao-2005}. The
relationships among the response data are difficult to explore when the
number of responses is large, and hence one might be inclined to think
that it could be helpful and more natural to consider the response as
a smooth real function. Moreover, with the rapid development of
accurate and sensitive instruments and thanks to the currently
available large storage resources, data are now often collected
in the form of curves or images. The statistical framework underlying
the analysis of these data as a single function observation rather
than a collection of individual observations is called functional data
analysis (FDA) and was first introduced by \citet{Ramsay-1991}.

It should be pointed out that in earlier studies a similar but less
explicit statement of the functional approach was addressed in
\citet{Dauxois-1982}, while the first discussion of what is meant by
``functional data'' appears to be by \citet{Ramsay-1982}. Functional
data analysis deals with the statistical description and modeling of
random functions. For a wide range of statistical tools, ranging from
exploratory and descriptive data analysis to linear models and
multivariate techniques, a functional version has been recently
developed.  Reviews of theoretical concepts and prospective
applications of functional data can be found in the two monographs by
\citet{Ramsay-2005, Ramsay-2002}.  One of the most crucial questions
related to this field is ``What is the correct way to handle large
data? Multivariate or Functional?"  Answering this question requires
better understanding of complex data structures and relationship among
variables. Until now, arguments for and against the use of a functional
data approach have been based on methodological considerations or
experimental investigations \citep{ferraty-2002, Rice-2004}. However,
we believe that without further improvements in theoretical issues
and in algorithm design of functional approaches, exhaustive comparative
studies will remain hard to conduct.

This motivates the general framework we develop in this paper. To the
best of our knowledge, nonlinear methods for functional data is a
topic that has not been sufficiently addressed in the FDA
literature. Unlike previous studies on nonlinear supervised
classification or real response regression of functional 
data~\citep{Rossi-2006, Ferraty-2004, Preda-2007}, this paper addresses the
problem of learning tasks where the output variables are functions. 
From a machine learning point of view, the problem can be viewed as
that of learning a function-valued function $f:\mathcal{X}
\longrightarrow\mathcal{Y}$ where $\mathcal{X}$ is the input space and 
$\mathcal{Y}$ the~(possibly infinite-dimensional) Hilbert space of the functional output data. Various situations can be
distinguished according to the nature of input data attributes 
(scalars or/and functions).  
We focus in this work on the case where input attributes are functions, too,
but it should be noted that the framework developed here can also be applied
when the input data are either discrete, or continuous. Lots of practical
applications involve a blend of both functional and non functional attributes, but 
we do not mix non functional attributes with
functional attributes in this paper. This point has been discussed in
\citep{Kadri-2011b}. To deal with non-linearity, we adopt a
kernel-based approach and we design operator-valued kernels that
perform the mapping between the two spaces of functions. Our main
results demonstrate how basic concepts and properties of kernel-based
learning known in the case of multivariate data can be restated for
functional data.

Extending learning methods from multivariate to functional response
data may lead to further progress in several practical problems of
machine learning and applied statistics. To compare the proposed
nonlinear functional approach with other multivariate or functional
methods and to apply it in a real world setting, we are interested in
the problems of speech inversion and sound recognition, which have
attracted increasing attention in the speech processing community in
the recent years~\citep{Mitra-2010,Asma-2008}.  These problems can be
cast as a supervised learning problem which include some components~(predictors or responses) that may be viewed as random curves. In this
context, though some concepts on the use of RKHS for functional data
similar to those presented in this work can be found
in~\citet{Lian-2007}, the present
paper provides a much more complete view of learning from functional
data using kernel methods, with extended theoretical analysis and
several additional experimental results.

This paper is a combined and expanded version of our previous conference 
papers~\citep{Kadri-2010,Kadri-2011c}. It gives the full justification, 
additional insights as well as new and comprehensive experiments that
strengthen the results of these preliminary conference papers. 
The outline of the paper is as follows. 
In Section~\ref{sec:FDA-ML}, we discuss the connection between the two fields Functional Data Analysis and Machine Learning, and outline our main contributions. Section~\ref{sec:notation} defines the notation used throughout the paper. Section~\ref{sec:frkhs}, 
describes the theory of reproducing kernel Hilbert spaces of
function-valued functions and  shows how vector-valued RKHS concepts
can be extended to infinite-dimensional output spaces.  In
Section~\ref{fk}, we exhibit a class of operator-valued kernels that
perform the mapping between two spaces of functions and discuss some
ideas for understanding their associated feature maps.  In
Section~\ref{fvfe}, 
we  provide a
function-valued function estimation procedure based on inverting block
operator kernel matrices, propose a learning algorithm that can handle functional data, and analyze theoretically its generalization properties. Finally in Section~\ref{sec:exp}, we
illustrate the performance of our approach through speech and audio processing
experiments.


\section{The Interplay of FDA and ML Research}
\label{sec:FDA-ML}

To put our work in context, we begin by discussing the interaction between functional data analysis~(FDA) and machine learning~(ML). Then, we give an overview of our contributions.

\medskip
Starting from the fact that ``new types of data require new tools for analysis'', FDA emerges as a well-defined and suitable concept to further improve classical multivariate statistical methods when data are functions~\citep{Levitin-2007}. This research field is currently very active, and considerable progress has been made in recent years in designing statistical tools for infinite-dimensional data that can be represented by real-valued functions rather than by discrete, finite dimensional vectors~\citep{Ramsay-2005, Ferraty-2006, Shi-2011, Horvath-2012}.
While the FDA viewpoint is conventionally adopted in the mathematical statistics community to deal with data in infinite-dimensional spaces, it does not appear to be commonplace for machine learners. One possible reason for this lack of success is that the formal use of infinite dimensional spaces for practical ML applications may seem unjustified; because in practice traditional measurement devices are limited in providing discrete and not functional data, and a machine learning algorithm can process only finitely represented objects. We believe that for applied machine learners it should be vital to know the full range of applicability of functional data analysis and infinite-dimensional data representations. But due to limitation of space we shall say only few words about the occurrence of functional data in real applications and about the real learning task lying behind this kind of approach. The reader is referred to~\citet{Ramsay-2002} for more details and references. Areas of application discussed and cited there include medical diagnosis, economics, meteorology, biomechanics, and education. For almost all these applications, the high-sampling rate of today's acquisition devices makes it natural to directly handle functions/curves instead of discretized data. Classical multivariate statistical methods may be applied to such data, but they cannot take advantage
of the additional information implied by the smoothness of the underlying functions. FDA methods can have beneficial effects in this direction by extracting additional information contained in the functions and their derivatives, not normally available through traditional methods~\citep{Levitin-2007}.


\begin{figure}[t]
\centering
\includegraphics[scale=0.5]{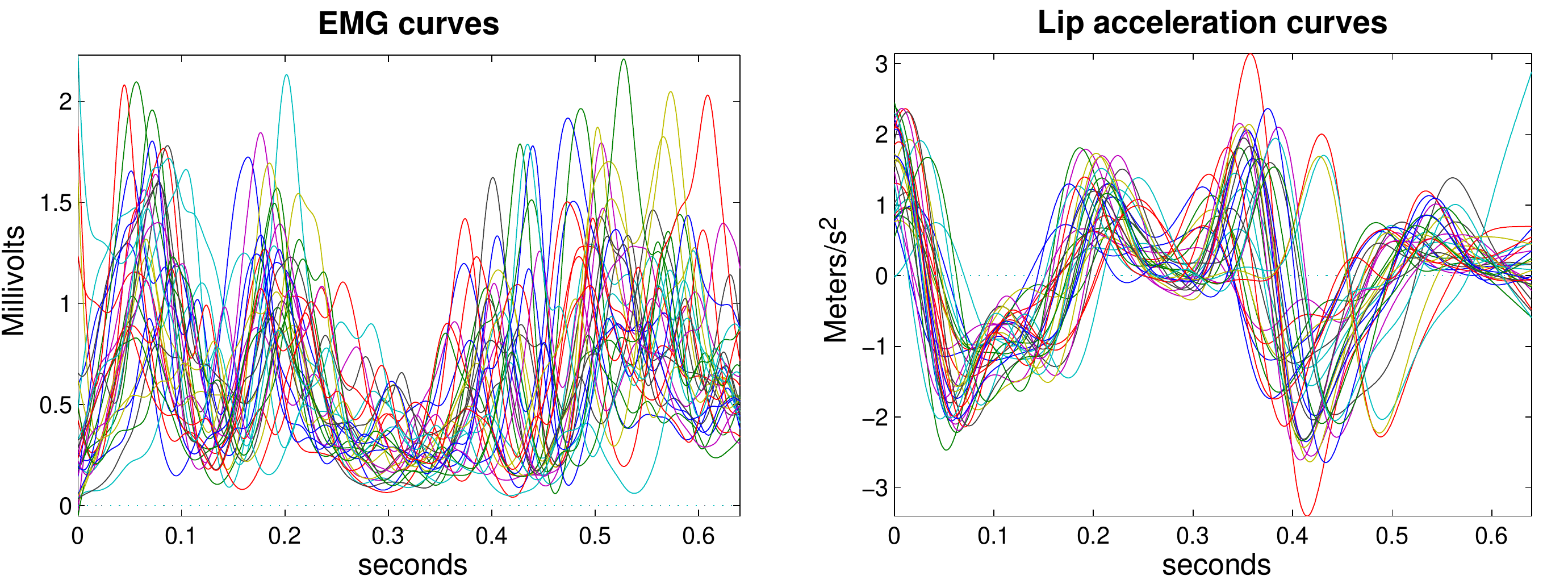}
\caption{Electromyography~(EMG) and lip acceleration curves. The left panel displays EMG recordings from a facial muscle that depresses the lower lip, the depressor labii inferior. The right panel shows 
the accelerations of the center of the lower lip of a speaker pronouncing the syllable ``bob'', embedded in the phrase ``Say bob again'', for 32 replications~\citep[Chapter~10]{Ramsay-2002}.\vspace{-0.2cm}}
\label{EMG_lip}
\end{figure}

\begin{figure}[t]
\centering
\includegraphics[scale=0.37]{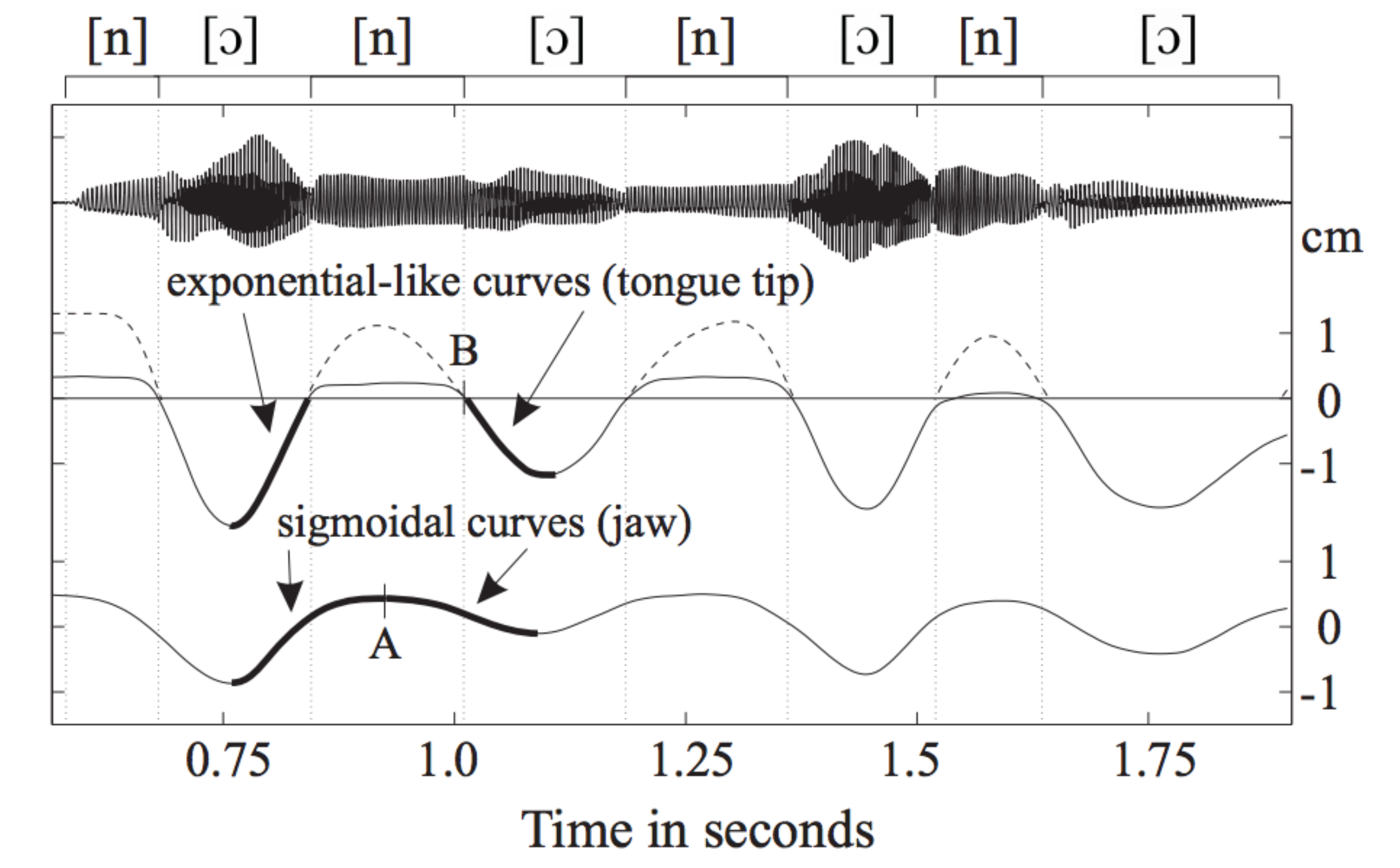}
\caption{
``Audio signal (top), tongue tip trajectory (middle), and jaw trajectory (bottom) for the utterance~[\textsecstress n\textipa{\textopeno}n\textipa{\textopeno}\textprimstress n\textipa{\textopeno}n\textipa{\textopeno}]. The trajectories were measured by electromagnetic articulography~(EMA) for coils on the tongue tip and the lower incisors. Each trajectory shows the displacement along the first principal component of the original two-dimensional trajectory in the midsagittal plane. The dashed curves show hypothetical continuations of the tongue tip trajectory towards and away from virtual targets during the closure intervals."~\citep{Birkholz-2010}.}
\label{PreviewSI}
\end{figure}

To get a better idea about the natural occurrence of functional data in ML tasks, Figure~\ref{EMG_lip} depicts a functional data set introduced by~\citet{Ramsay-2002}. The data set consists of 32 records of the movement of the center of the lower lip when a subject was repeatedly required to say the syllable ``bob'', embedded in the sentence, ``Say bob again'' and the corresponding EMG activities of the primary muscle depressing the lower lip, the depressor labii inferior (DLI)\footnote{The data set is available at \url{http://www.stats.ox.ac.uk/~silverma/fdacasebook/lipemg.html}. More information about the data
collection process can be found in~\citet[Chapter~10]{Ramsay-2002}.}. The goal here is to study the dependence of the acceleration of 
the lower lip in speech on neural activity.
EMG and lip accelerations curves can be well modeled by continuous functions of time that allow to capture functional dependencies and interactions between samples~(feature values). Thus, we face a regression problem where both input and output data are functions.
In much the same way, Figure~\ref{PreviewSI} also shows a ``natural'' representation of data in terms of functions~\footnote{This figure is from~\citet{Birkholz-2010}.}. It represents a speech signal used for acoustic-articulatory speech inversion and  produced by a subject pronouncing a sequence of~[\textsecstress CVCV\textprimstress CVCV]~(C=consonant, V=vowel) by combining the vowel $\{/ \text{\textipa{\textopeno}} /\}$ with the consonant $\{/\text{n}/\}$. 
The articulatory trajectories are represented by the upper and lower solid curves that show the displacement of fleshpoints on the tongue tip and the jaw along the main movement direction of these points during the repeated opening and closing gestures. This example is from a recent study on the articulatory modeling of speech signals~\citep{Birkholz-2010}. The concept of articulatory gestures in the context of speech-to-articulatory inversion will be explained in more details in Section~\ref{sec:exp}. As shown in the figure, the observed articulatory trajectories are typically modeled by smooth functions of time with periodicity properties and exponential or sigmoidal shape, and the goal of speech inversion is to predict and recover geometric data of the vocal tract from the speech information.

In both examples given above, response data clearly present a functional behavior that should be taken into account during the learning process. We think that handling these data as what they really are, that is functions, is a promising way to tackle prediction problems and design efficient ML systems for continuous data variables. 
Moreover, ML methods which can handle functional features can open up plenty of new areas of application, where the flexibility of functional and infinite-dimensional spaces would allow to enable us to achieve significantly better performance while managing huge amounts of training data.

In the light of these observations, there is an interest in overcoming methodological and practical problems that hinder the wide adoption and use of functional methods built for infinite-dimensional data. Regarding the practical issue related to the application and implementation of infinite-dimensional spaces, a standard means of addressing it is to choose a functional space \textit{a priori} with a known predefined set of basis functions in which the data will be mapped. This may include a preprocessing step, which consists in converting the discretized data into functional objects using interpolation or approximation techniques. Following this scheme, parametric FDA methods have emerged as a common approach to extend multivariate statistical analysis in functional and infinite-dimensional situations~\citep{Ramsay-2005}. More recently, nonparametric FDA methods have received increasing attention because of their ability to avoid fixing a set of basis functions for the functional data beforehand~\citep{Ferraty-2006}. These methods are based on the concept of semi-metrics for modeling functional data. 
The reason for using a semi-metric rather than a metric is that the coincidence axiom, namely $d(x_i, x_j) = 0 \Leftrightarrow x_i=x_j$, may result in curves with very similar shapes being categorized as  
distant~(not similar to each other). To define closeness between functions in terms of shape rather than location semi-metrics can be used.
In this spirit, \citet{Ferraty-2006} provided a semi-metric based methodology for nonparametric functional data analysis and argued that this can be a sufficiently general theoretical framework to tackle infinite-dimensional data without being “too heavy” in terms of computational time and implementation complexity.

\medskip
Thus, although both parametric and nonparametric functional data analyses deal with infinite-dimensional data, they are computationally feasible and quite practical since the observed functional data are approximated in a basis of the function space with possibly finite number of elements. What we really need is the inner or semi-inner product of the basis elements and the representation of the functions with respect to that basis.
We think that Machine Learning research can profit from exploring other representation formalisms that support the expressive power of functional data. Machine learning methods which can accommodate functional data should open up new possibilities for handling practical applications for which the flexibility of infinite-dimensional spaces could be exploited to achieve performance benefits and accuracy gains.
On the other hand, in the FDA field, there is clearly a need for further development of computationally efficient and understandable algorithms that can deliver near-optimal solutions for infinite-dimensional problems and that can handle a large number of features. The transition from infinite-dimensional statistics to efficient algorithmic design and implementation is of central importance to FDA methods in order to make them more practical and popular. In this sense, Machine Learning can have a profound impact on FDA research.

 \medskip
In reality, ML and FDA have more in common than it might seem. 
There are already existing machine learning algorithms that can also be viewed as FDA methods. For example, these include kernel methods which use a certain type of similarity measure (called a kernel) to map observed data in a high dimensional feature space, in which linear methods are used for learning problems~\citep{Shawetaylor-2004, Scholkopf-2002}. Depending on the choice of the kernel function, the feature space can be infinite-dimensional. The kernel trick is used, allowing to work with finite Gram matrix of inner products between the possibly infinite-dimensional features which can be seen as functional data.  This connection between kernel and FDA methods is clearer with the concept of kernel embedding of probability distributions, where, instead of (observed) single points, kernel means are used to represent probability distributions~\citep{Smola-2007,Sri-2010}. The kernel mean corresponds to a mapping of a probability distribution in a feature space which is rich enough so that its expectation uniquely identifies the distribution. Thus, rather than relying on large collections of vector data, kernel-based learning can be adapted to probability distributions that are constructed to meaningfully represent the discrete data by the use of kernel means~\citep{Muandet-2012}. In some sense, this represents 
a similar design to FDA methods, where data are assumed to lie in a functional space even though they are acquired in a discrete manner. 
There are also other papers that deal with machine learning problems where covariates are probability distributions and discuss their relation with FDA~\citep{Poczos-2012,Poczos-2013,Oliva-2013}.
At that point, however, the connection between ML and FDA is admittedly weak and needs to be bolstered by the delivery of more powerful and flexible learning machines that are able to deal with functional data and infinite-dimensional spaces.

\medskip
In the FDA field, linear models have been explored extensively. Nonlinear modeling of functional data is, however, a topic that has not been sufficiently investigated, especially when response data are functions. Reproducing kernels provide a powerful tool for solving learning problems with nonlinear models, but to date they have been used more to learn scalar-valued or vector-valued functions than function-valued functions. Consequently, kernels for functional response data and their associated function-valued reproducing kernel Hilbert spaces have remained mostly unknown and poorly studied. In this work, we aim to rectify this 
situation, and highlight areas of overlap between the two fields FDA and ML, particularly with regards to the applicability and relevance of the FDA paradigm coupled with machine learning techniques. 
Specifically, we provide a learning methodology for nonlinear FDA based on the theory of reproducing kernels. 
The main contributions are as follows: 
\begin{itemize}
\item we introduce a set of rigorously defined operator-valued kernels suitable for functional response data, that can be valuably applied to model dependencies between samples and take into account the functional nature of the data, like the smoothness of the curves underlying the discrete observations,
\item we propose an efficient algorithm for learning function-valued functions~(operators) based on the spectral decomposition of block operator matrices,
\item we study the generalization performance of our learned nonlinear FDA model using the notion of algorithmic stability,
\item we show the applicability and suitability of our framework to two problems in audio signal processing, namely speech inversion and sound recognition, where features are functions that are dependent on each other. 
\end{itemize}



\section{Notations and Conventions} 
\label{sec:notation}

We start by some standard notations and definitions used all along the paper. 
Given a Hilbert space $\mathcal{H}$, $\langle \cdot,\cdot \rangle_\mathcal{H}$ and 
$\| \cdot \|_\mathcal{H}$ refer to its inner product and norm, respectively.
{$H^n = \underbrace{\mathcal{H}\times\ldots\times\mathcal{H}}_{n \mbox{\scriptsize{} times}}$, $n \in \mathbb{N}_+$, denotes the topological product of $n$ spaces $\mathcal{H}$.}
We denote by $\mathcal{X} = \{x: \Omega_{x} \longrightarrow \mathbb{R} \}$ and 
$\mathcal{Y} = \{y: \Omega_{y} \longrightarrow \mathbb{R} \}$ the separable Hilbert spaces of input and 
output real-valued functions whose domains are $\Omega_{x}$ and $\Omega_{y}$, respectively.
In functional data analysis 
domain, the space of functions is generally assumed to be the Hilbert space of equivalence classes of square integrable functions, denoted by $L^2$. Thus, in the rest of the paper, 
we consider $\mathcal{Y}$ to be the space $L^2(\Omega_{y})$, where $\Omega_{y}$ is a compact set.
The vector space of functions from $\mathcal{X}$ into $\mathcal{Y}$ is denoted by $\mathcal{Y}^\mathcal{X}$ endowed with the topology of uniform convergence on compact subsets of $\mathcal{X}$.
We denote by $\mathcal{C(X,Y)}$ the vector space of continuous functions from $\mathcal{X}$ to $\mathcal{Y}$, by $\mathcal{F}\subset \mathcal{Y}^\mathcal{X}$ the Hilbert space of function-valued functions $F: \mathcal{X} \longrightarrow \mathcal{Y}$, 
 and by $\mathcal{L(Y)}$ the set of bounded linear operators from $\mathcal{Y}$ to $\mathcal{Y}$.

{We now fix the following conventions for bounded linear operators and block operator matrices.}
\begin{de}(adjoint, self-adjoint, and positive operators) \vspace{1mm} \\ 
Let $A \in \mathcal{L(Y)}$. Then:
 \begin{enumerate}[\normalfont (i)]
    \item $A^*$, the adjoint operator of $A$, is the unique operator in $\mathcal{L(Y)}$ that satisfies 
\begin{equation*}
 \langle A y, z \rangle_{\mathcal{Y}}  = \langle  y, A^* z \rangle_{\mathcal{Y}} \ ,    \ \forall y \in \mathcal{Y} , \forall z \in \mathcal{Y},                                                                                                                                                                                                         
\end{equation*}
\item $A$ is self-adjoint if $A = A^*$,

\item $A$ is positive if it is self-adjoint and $\forall y \in \mathcal{Y}$, 
$\langle A y, y \rangle_{\mathcal{Y}}   \geq 0 $~(we write $A\geq 0$),

\item $A$ is larger or equal than $B\in \mathcal{L(Y)}$, if $A-B$ is positive, 
\textit{i.e.},  $\forall y \in \mathcal{Y}$, $\langle A y, y \rangle_{\mathcal{Y}}   \geq \langle B y, y \rangle_{\mathcal{Y}}$~(we write $A\geq B$).
\end{enumerate}
 \label{operator}
\end{de}

\begin{de}(block operator matrix) \vspace{1mm} \\ 
Let $n \in \mathbb{N}$, let $\mathcal{Y}^n = \underbrace{\mathcal{Y} \times \ldots \times \mathcal{Y}}_{n \mbox{\scriptsize{} times}}$. 

\begin{enumerate}[\normalfont (i)]
    \item
$\mathbf{A} \in \mathcal{L}(\mathcal{Y}^n)$, 
given by
\begin{equation*}
\nonumber
 \mathbf{A} = 
\begin{pmatrix}
A_{11} & \ldots & A_{1n} \\
\vdots &  & \vdots \\
A_{n1} & \ldots & A_{nn} \\
\end{pmatrix} 
\end{equation*}
where each $A_{ij}\in \mathcal{L(Y)}$, $i,j = 1,\ldots,n$, is called a block operator matrix,
 
\item \label{enumblockoperator} the adjoint (or transpose) of $\mathbf{A}$ is the block operator matrix 
$\mathbf{A^*} \in \mathcal{L}(\mathcal{Y}^n)$ such that $(A^*)_{ij}= (A_{ji})^*$,

\item self-adjoint and order relations of block operator matrices are defined in 
the same way as for bounded operators (see Definition~\ref{operator}).
\end{enumerate}
\label{blockoperator}
\end{de}
%
Note that item~(\ref{enumblockoperator}) in Definition~\ref{blockoperator} is obtained from the definition of  adjoint operator. 
It is easy to see that $\forall \mathbf{y} \in \mathcal{Y}^n$ and $\forall \mathbf{z} \in \mathcal{Y}^n$; we have: 
$\langle  \mathbf{A} \mathbf{y},  \mathbf{z} \rangle_{\mathcal{Y}^n} = \sum\limits_{i,j} \langle A_{ij} y_j,  z_i \rangle_{\mathcal{Y}}
 = \sum\limits_{i,j} \langle   y_j, A_{ij}^* z_i \rangle_{\mathcal{Y}} = \sum\limits_{i,j} \langle   y_j, (A^*)_{ji} z_i \rangle_{\mathcal{Y}} 
= \langle   \mathbf{y},  \mathbf{A}^* \mathbf{z} \rangle_{\mathcal{Y}^n} $, where $(A^*)_{ji} = (A_{ij})^*$.
\\[0.2cm]

To help the reader, notations frequently used in the paper are summarized in Table~\ref{tab:notation}. 

\begin{table}[t]
 \centering
\begin{tabular}{|r|c|l|}
\hline
 real numbers &  $\alpha$, $\beta$, $\gamma$, $\ldots$ & Greek characters \\
 integers &  $i$, $j$, $m$, $n$ & \\
vector spaces\footnotemark & $\mathcal{X}$, $\mathcal{Y}$, $\mathcal{H}$, $\ldots$ & Calligraphic letters \\ 
subsets of the real plain & $\Omega$, $\Lambda$, $\Gamma$, $\ldots$ & capital Greek characters \\
 functions\footnotemark (or vectors) & $x$, $y$, $f$, $\ldots$ & small Latin characters \\
vector of functions & $\mathbf{u}$, $\mathbf{v}$, $\mathbf{w}$, $\ldots$ & small bold Latin characters \\
operators (or matrices) & $A$, $B$, $K$, $\ldots$ & capital Latin characters \\
block operator matrices & $\mathbf{A}$, $\mathbf{B}$, $\mathbf{K}$, $\ldots$ & capital bold Latin characters \\
adjoint operator & $*$ & $A^*$ adjoint of operator $A$ \\
identical equality & $\equiv$ &  equality of mappings\\
 definition & $\triangleq$ &  equality by definition\\
\hline
\end{tabular}
\caption{Notations used in this paper.}
\label{tab:notation}
\end{table}
\addtocounter{footnote}{-1}
\footnotetext{We also use the standard notations such as $\mathbb{R}^n$ and $L^2$.}
\addtocounter{footnote}{+1}
\footnotetext{We denote by small Latin characters scalar-valued functions. 
Operator-valued functions (or kernels) are denoted by capital Latin characters  
$A(\cdot,\cdot)$, $B(\cdot,\cdot)$, $K(\cdot,\cdot)$, $\ldots$}


\section{Reproducing Kernel Hilbert Spaces of Function-valued Functions}
\label{sec:frkhs}

Hilbert spaces of scalar-valued functions with reproducing kernels were
introduced and studied in~\citet{Aronszajn-1950}. Due to their crucial
role in designing kernel-based learning methods, these spaces have
received considerable attention over the last two 
decades~\citep{Shawetaylor-2004, Scholkopf-2002}. 
More recently, interest has
grown in exploring reproducing Hilbert spaces of vector functions for learning
vector-valued functions~\citep{Micchelli-2005a, Carmeli-2006, Caponnetto-2008, Carmeli-2010, Zhang-2012}, even
though the idea of extending the theory of Reproducing Kernel Hilbert
Spaces from the scalar-valued case to the vector-valued one is not new and
dates back to at least~\citet{Schwartz-1964}. For more details, see the review paper by~\citet{Alvarez-2012}.

In the field of machine learning, \citet{Evgeniou-2005} have shown how Hilbert spaces of vector-valued functions and matrix-valued reproducing kernels can be used in the context of multi-task learning, with the goal of learning many related
regression or classification tasks simultaneously. %
Since this seminal work, it has been demonstrated that these kernels and their associated spaces are capable of solving various other learning problems such as multiple output learning~\citep{Baldassarre-2012}, manifold regularization~\citep{Minh-2011}, structured output prediction~\citep{Brouard-2011,Kadri-2013mvk}, multi-view learning~\citep{Minh-2013, Kadri-2013} and network inference~\citep{Lim-2013,Lim-2014}. %

In contrast to most of these previous works, here we are interested in the general case where the output space is a space of vectors with infinite dimension. This may be valuable from a variety of perspectives. Our main motivation is the supervised learning problem when output data are functions that could represent, for example, one-dimensional curves~(this was mentioned as future work in~\citealt{Szedmak-2006}).  One of the simplest ways to handle these data is to treat them as multivariate vectors. However this method does not
consider any dependency of different values over subsequent time-points within the same functional datum and suffers when data dimension is very large. Therefore, we adopt a functional data analysis viewpoint~\citep{Zhao-2004, Ramsay-2005,Ferraty-2006} in which multiple curves are viewed as functional realizations of a single function.
It is important to note that matrix-valued kernels for infinite-dimensional output spaces, commonly known as operator-valued kernels,  have been considered in previous studies~\citep{Micchelli-2005a, Caponnetto-2008, Carmeli-2006, Carmeli-2010}; however, they have been only studied in a theoretical perspective. Clearly, further  investigations are needed to illustrate  the practical benefits of the use of operator-valued kernels, which is the main focus of this work.

\medskip
We now describe how RKHS theory can be extended from real or vector to
functional response data.  In particular, we focus on reproducing
kernel Hilbert spaces whose elements are function-valued functions~(or operators) and
we demonstrate how basic properties of real-valued RKHS can be restated in
the functional case, if appropriate conditions are satisfied. Extension to the functional case is not so
obvious and requires tools from functional
analysis~\citep{Rudin-1991}.
Spaces of operators whose range is infinite-dimensional can exhibit unusual 
behavior, and standard topological properties may not be preserved in the infinite-dimensional case because of functional analysis subtleties. So, additional restrictions imposed on these spaces are needed for extending the theory of RKHS towards infinite-dimensional output spaces.
Following~\citet{Carmeli-2010}, we mainly focus on separable Hilbert spaces with reproducing operator-valued kernels whose elements are continuous functions.
This is a sufficient condition to avoid %
topological and measurability problems %
encountered with this extension.
For more details about vector or function-valued RKHS of measurable and continuous functions, see~\citet[Sections 3 and 5]{Carmeli-2006}.
Note that the framework developed in this section should be valid for any type of input
data~(vectors, functions, or structures).  In this paper, however, we consider the case
where both input and output data are functions. 
\begin{de} (Operator-valued kernel) \vspace{1mm} \\ 
  An $\mathcal{L(Y)}$-valued kernel $K$ on
  $\mathcal{X}^2$ is a function
  $K(\cdot,\cdot):\mathcal{X} \times \mathcal{X}
  \longrightarrow \mathcal{L(Y)}$;
   \begin{enumerate}[\normalfont (i)]
   \itemsep=-0.05cm
    \item $K$ is Hermitian if $\forall x,z\in\mathcal{X}$,
      $K(w,z)=K(z,w)^{*}$, 
      (where the superscript * denotes the adjoint operator), 
    \item $K$ is nonnegative on $\mathcal{X}$ if it
      is Hermitian 
      and for every natural number $r$ and all
      $\{(w_{i},u_{i})_{i=1,\ldots ,r}\}\in \mathcal{X} \times
      \mathcal{Y}$, the matrix with ij-th entry
      $\langle K(w_{i},w_{j})u_{i},u_{j}\rangle _{\mathcal{Y}}$
      is nonnegative~(positive-definite).
   \end{enumerate}
\end{de}

\begin{de} (Block operator kernel matrix) \vspace{1mm} \\ 
  Given a set $\{w_i\} \in \x$, $i=1,\ldots,n$ with $n\in\mathbb{N}_+$, and an operator-valued kernel $K$, the corresponding block operator kernel matrix is the matrix $\mathbf{K} \in \mathcal{L}(\mathcal{Y}^n)$ with entries
  $$ \mathbf{K}_{ij} = K(w_i,w_j).$$
  \end{de}
 The block operator kernel matrix is simply the kernel matrix associated to an operator-valued kernel. Since the kernel outputs an operator, the kernel matrix is in this case a block matrix where each block is an operator in $\mathcal{L(Y)}$.
 It is easy to see that an operator-valued kernel $K$ is nonnegative if and only if the associated block operator kernel matrix $\mathbf{K}$ is positive.

\begin{de} (Function-valued RKHS) \vspace{1mm} \\ 
  A Hilbert space $\mathcal{F}$ of functions from $\mathcal{X}$ to
  $\mathcal{Y}$ is called a reproducing kernel Hilbert space if
  there is a nonnegative $\mathcal{L(Y)}$-valued kernel
  $K$ on $\mathcal{X}^2$ such that:
  \begin{enumerate}[\normalfont (i)]
    \item \label{enum1:i} the function $z \longmapsto
      K(w,z)g$ belongs to $\mathcal{F}, \forall z, w \in
      \mathcal{X}$ and $g \in \mathcal{Y}$, 
    \item \label{enum1:ii} for every $F \in \mathcal{F}, w\in\mathcal{X}$ and $g \in \mathcal{Y}$,
      $\langle F,K(w,\cdot)g\rangle _{\mathcal{F}} =
      \langle F(w),g\rangle _{\mathcal{Y}}$.
  \end{enumerate}
\end{de}
On account of (\ref{enum1:ii}), the kernel is called the reproducing
kernel of $\mathcal{F}$.
In~\citet[Section~5]{Carmeli-2006}, the authors provided a characterization of RKHS with operator-valued kernels whose functions are continuous and  proved that $\mathcal{F}$ is a subspace of $\mathcal{C(X,Y)}$, the vector space of continuous functions from $\x$ to $\y$,  if and only if the reproducing kernel $K$ is locally bounded and separately continuous. Such a kernel is qualified as Mercer~\citep{Carmeli-2010}. In the following, we will only consider separable RKHS $\mathcal{F} \subset \mathcal{C(X,Y)}$. 
\begin{theo} (Uniqueness of the reproducing operator-valued kernel) \vspace{1mm} \\
  \label{th:unicityOfTheReproducingKernel}
  If a Hilbert space $\mathcal{F}$ of functions from $\mathcal{X}$ to $\mathcal{Y}$ 
admits a reproducing kernel, then the reproducing kernel
  $K$ is uniquely determined by the Hilbert space
  $\mathcal{F}$.
\end{theo}
%
%
\textbf{Proof}: Let $K$ be a reproducing kernel of
$\mathcal{F}$. Suppose that there exists another reproducing kernel
$K'$ of $\mathcal{F}$. Then, for all $\{w, w'\} \in \mathcal{X}$ and $\{h,g\} \in \mathcal{Y}$, applying the reproducing property for $K$
and $K'$ we get 
\begin{equation}
\label{th1-1}
  \langle K'(w',\cdot)h,K(w,\cdot)g\rangle _{\mathcal{F}} = \langle K'(w',w)h,g\rangle _{\mathcal{Y}},
\end{equation}
we have also
\begin{align}
\label{th1-2}
\begin{array}{ccccc}
  \langle K'(w',\cdot)h,K(w,\cdot)g\rangle_{\mathcal{F}} 
  & =&\langle K(w,\cdot)g,K'(w',\cdot)h\rangle _{\mathcal{F}} 
  &=&\langle K(w,w')g,h\rangle _{\mathcal{Y}} \\
  &=&\langle g,K(w,w')^{*}h\rangle _{\mathcal{Y}} 
  &=&\langle g,K(w',w)h\rangle _{\mathcal{Y}}. 
\end{array}
\end{align}
(\ref{th1-1}) and~(\ref{th1-2}) $\Rightarrow K(w,w')\equiv
K'(w,w')$, $\forall w, w' \in \mathcal{X}$.
\hfill $\blacksquare$ 

%
\bigskip

A key point for learning with kernels is the ability to express
functions in terms of a kernel providing the way to evaluate a
function at a given point. This is possible because there exists a
bijection relationship between a large class of kernels and associated
reproducing kernel spaces which satisfy a regularity property.
Bijection between scalar-valued kernels and RKHS was first established
by~\citet[Part~I, Sections~3~and~4]{Aronszajn-1950}. Then~\citet[Chapter~5]{Schwartz-1964} shows that this
is a particular case of a more general situation.  
This bijection in the case where input and output data are continuous
and belong to the infinite-dimensional functional spaces $\mathcal{X}$ and $\mathcal{Y}$, respectively, is still valid and is given by the following theorem~(see also Theorem~4 of~\citealp{Senkene-1973}).

\begin{theo} (Bijection between function-valued RKHS and operator-valued kernel) \vspace{1mm}  \\
  \label{th:positiveDefiniteImpliesKernel}
  A $\mathcal{L(Y)}$-valued {Mercer} kernel
  $K$ on $\mathcal{X}^2$ is the reproducing
  kernel of some Hilbert space $\mathcal{F}$, if and only if it is nonnegative.
\end{theo}
We give a proof of this theorem by extending the scalar-valued case
$\mathcal{Y} = \mathbb{R}$ in~\citet{Aronszajn-1950} to the domain
of functional data analysis domain where $\mathcal{Y}$ is $L^2(\Omega_y)$.\footnote{The proof should be applicable to arbitrarily separable output Hilbert spaces $\mathcal{Y}$.}   The proof is performed in two
steps. The necessity is an immediate result from the reproducing
property. For the sufficiency, the outline of the proof is as follows:
we assume $\mathcal{F}_{0}$ to be the space of all
$\mathcal{Y}$-valued functions $F$ of the form
$F(\cdot)=\sum_{i=1}^{n}K(w_{i},\cdot)u_{i}$, where
$w_{i} \in \mathcal{X}$ and $u_{i} \in \mathcal{Y}$,
with the following inner product 
$\left\langle F(\cdot),G(\cdot)\right\rangle_{\mathcal{F}_{0}}  =  
\sum_{i=1}^{n}\sum_{j=1}^{m}\left\langle K(w_{i},z_{j})u_{i},v_{j}\right\rangle_{\mathcal{Y}}$ defined for any $G(\cdot)=\sum_{j=1}^{m}K(z_{j},\cdot)v_{j}$ with $z_j\in \mathcal{X} $ and $v_j \in \mathcal{Y}$. 
We show that~$(\mathcal{F}_{0},\langle\cdot,\cdot\rangle_{\mathcal{F}_{0}})$ is
a pre-Hilbert space. Then we complete this pre-Hilbert space via Cauchy sequences~$\{F_{n}(\cdot)\} \subset \mathcal{F}_{0}$ to construct the Hilbert space
$\mathcal{F}$ of $\mathcal{Y}$-valued functions. Finally, we conclude that $\mathcal{F}$ is a reproducing kernel
Hilbert space, since $\mathcal{F}$ is a real inner product space that
is complete under the norm $\|\cdot\|_{\mathcal{F}}$ defined by 
$\|F(\cdot)\|_{\mathcal{F}}=\lim\limits_{n\rightarrow\infty}\|F_{n}(\cdot)\|_{\mathcal{F}_{0}} $, and
has $K(\cdot,\cdot)$ as reproducing kernel.
%
\\
\\
\textbf{Proof}:
\textit{Necessity}. Let $K$ be the reproducing kernel of a Hilbert
space $\mathcal{F}$. Using the reproducing property of the kernel
$K$ we obtain for any $\{w_{i}, w_{j}\}\in \mathcal{X}$ and $\{u_{i}, u_{j}\}\in
\mathcal{Y}$
\begin{eqnarray*}
  \begin{array}{l}
    \displaystyle\sum\limits_{i,j=1}^{n}\langle K(w_{i},w_{j})u_i,u_{j}\rangle _{Y} =
    \sum\limits_{i,j=1}^{n}\langle K(w_{i},\cdot)u_{i},K(w_{j},\cdot)u_{j}\rangle _{\mathcal{F}} \\
    =\langle \displaystyle\sum\limits_{i=1}^{n}K(w_{i},\cdot)u_{i},\sum\limits_{i=1}^{n}K(w_{i},\cdot)u_{i}\rangle _{\mathcal{F}}
    =\|\sum\limits_{i=1}^{n}K(w_{i},\cdot)u_{i}\|_{\mathcal{F}}^{2}\
    \geq \ 0.
  \end{array}
\end{eqnarray*}
\newline
\textit{Sufficiency}. Let $\mathcal{F}_{0}\subset\mathcal{Y}^\mathcal{X}$ be the space of all
$\mathcal{Y}$-valued functions $F$ of the form
$F(\cdot)=\sum\limits_{i=1}^{n}K(w_{i},\cdot)u_{i}$, where
$w_{i} \in \mathcal{X}$ and $u_{i} \in \mathcal{Y}$,
$i=1,\ldots, n$. We define the inner product of the
functions $F(\cdot)=\sum\limits_{i=1}^{n}K(w_{i},\cdot)u_{i}$ and 
$G(\cdot)=\sum\limits_{j=1}^{m}K(z_{j},\cdot)v_{j}$ from $\mathcal{F}_{0}$ as follows
\begin{align*}
\langle F(\cdot),G(\cdot)\rangle _{\mathcal{F}_{0}}=\langle \sum\limits_{i=1}^{n}K(w_{i},\cdot)u_{i},\sum\limits_{j=1}^{m}K(z_{j},\cdot) 
v_{j}\rangle _{\mathcal{F}_{0}} 
 = \sum\limits_{i=1}^{n} \sum\limits_{j=1}^{m} \langle 
K(w_{i},z_{j})u_{i},v_{j}\rangle _{\mathcal{Y}}.
\end{align*}
$\langle F(\cdot),G(\cdot)\rangle _{\mathcal{F}_{0}}$ is a symmetric bilinear form on
$\mathcal{F}_{0}$ and due to the positivity of the kernel
$K$, $\|F(\cdot)\|$ defined by \begin{equation*}\|F(\cdot)\|=\sqrt{\langle F(\cdot),F(\cdot)\rangle _{\mathcal{F}_{0}}}\end{equation*} is
a quasi-norm in $\mathcal{F}_{0}$. 
The reproducing property in $\mathcal{F}_{0}$ is verified with the
kernel $K$. In fact, if $F\in \mathcal{F}_{0}$ then
\begin{equation*}
F(\cdot)=\sum\limits_{i=1}^{n}K(w_{i},\cdot)u_{i},
\end{equation*}
and $
\forall\ (w,u) \in \mathcal{X}\times \mathcal{Y}$,
\begin{equation*}
    \langle F ,K(w,\cdot) u \rangle_{\mathcal{F}_{0}}=\langle 
\sum\limits_{i=1}^{n}K(w_{i},\cdot)u_{i},K(w,\cdot) u \rangle_{\mathcal{F}_{0}} 
   =\langle \sum\limits_{i=1}^{n}K(w_{i},w) u_{i},u \rangle_{\mathcal{Y}}
    =\langle F(w), u \rangle _{\mathcal{Y}}.
\end{equation*}
Moreover using the Cauchy-Schwartz inequality, we have: $\forall\
(w, u) \in \mathcal{X}\times \mathcal{Y}$,
\begin{align*}
  \langle F(w), u \rangle _{\mathcal{Y}}=\langle F(\cdot),K(w,\cdot) u \rangle_{\mathcal{F}_{0}} 
  \leq
  \|F(\cdot)\|_{\mathcal{F}_{0}}\|K(w,\cdot) u \|_{\mathcal{F}_{0}}.
\end{align*}
Thus, if $\|F\|_{\mathcal{F}_{0}} = 0$, then
$\langle F(w), u \rangle _{\mathcal{Y}} = 0$ for any $w$ and $u$, and
hence $F\equiv 0$. Thus~$(\mathcal{F}_{0},\langle .,.\rangle _{\mathcal{F}_{0}})$ is
a pre-Hilbert space. This pre-Hilbert space is in general not
complete, but it can be completed via Cauchy sequences to build the $\mathcal{Y}$-valued Hilbert space $\mathcal{F}$ which has $K$ as reproducing kernel, which concludes the proof. The completion of $\mathcal{F}_0$ is given in Appendix~\ref{App:completion}~(we refer the reader to the monograph by~\citealp{Rudin-1991}, for more details about completeness and the general theory of topological vector spaces).
\hfill $\blacksquare$
%

%
\bigskip

We now give an example of a function-valued RKHS and its operator-valued kernel. 
This example serves to illustrate how these spaces and their associated kernels generalize the standard scalar-valued case or the vector-valued one to functional and infinite-dimensional output data.
Thus, we first report an example of a scalar-valued RKHS and the corresponding scalar-valued kernel. We then extend this example to the case of vector-valued Hilbert spaces with matrix-valued kernels, and finally to function-valued RKHS where the output space is infinite dimensional. For the sake of simplicity, the input space $\mathcal{X}$ in these examples is assumed to be a subset of $\mathbb{R}$.

\begin{example}(Scalar-valued RKHS and its scalar-valued kernel; see~\citet{Canu-2003}) \vspace{1mm} \\
\label{exp:sv} 
Let $\mathcal{F}$ be the space defined as follows:
\begin{equation*}
  \left\{
    \begin{aligned}
    & \mathcal{F} = \big\{f:[0,1]\longrightarrow \mathbb{R} \text{ absolutely continuous},\ \exists f'  \in {L}^2([0,1]), 
f(x) = \int_0^x f'(z) dz \big\},\\ 
   & \langle f_1, f_2 \rangle_\mathcal{H} = \langle f'_1, f'_2 \rangle_{{L}^2([0,1])}.
    \end{aligned}
  \right.
\end{equation*}
$\mathcal{F}$ is the Sobolev space of degree 1, also called the Cameron-Martin space, and is a scalar-valued RKHS of functions $f:[0,1]\longrightarrow \mathbb{R}$ with the scalar-valued reproducing kernel $k(x,z)=min(x,z)$, $\forall x,z \in \mathcal{X}=[0,1]$.
\end{example}

\begin{example}(Vector-valued RKHS and its matrix-valued kernel) \vspace{1mm} \\ 
\label{exp:vv}
Let $\mathcal{X}=[0,1]$ and $\mathcal{Y}=\mathbb{R}^n$. Consider the matrix-valued kernel $K$ defined by:
\begin{equation}
K(x,z)=
\left\lbrace
\begin{array}{cc}
diag(x)  & \mbox{if} \quad  x\leq z,\\
diag(z) & otherwise,\\
\end{array}\right.
\end{equation}
where, $\forall a\in\mathbb{R}$, $diag(a)$ is the $n\times n$ diagonal matrix with diagonal entries equal to $a$. 
Let $\mathcal{M}$ be
the space of vector-valued functions from $\mathcal{X}$ onto $\mathbb{R}^n$ whose norm 
$ \|g\|_{\mathcal{M}}^2 = \displaystyle\sum_{i=1}^n \int_\mathcal{X} [g(x)]_i^2 dx$ is finite.\\
The matrix-valued mapping $K$ is the reproducing kernel of the vector-valued RKHS $\mathcal{F}$ defined as follows:
\begin{equation*}
  \left\{
    \begin{aligned}
    & \mathcal{F} = \big\{f: [0,1]\longrightarrow \mathbb{R}^n,\ \exists f' = \displaystyle\frac{d f(x)}{d x} \in \mathcal{M}, 
[f(x)]_i = \int_0^x [f'(z)]_i dz, \forall i =1,\ldots,n \big\},\\ 
   & \langle f_1, f_2 \rangle_\mathcal{F} = \langle f'_1, f'_2 \rangle_{\mathcal{M}}.
    \end{aligned}
  \right.
\end{equation*}
Indeed, $K$ is nonnegative and we have, $\forall x \in \mathcal{X}$, $y\in \mathbb{R}^n$ and $f\in \mathcal{F}$,
\begin{align*}
\begin{array}{l}
  \langle f, K(x,\cdot)y \rangle_{\mathcal{F}} = \langle f', [K(x,\cdot)y]' \rangle_{\mathcal{M}}\\[0.1cm]
= \displaystyle\sum_{i=1}^n \int_0^1  [f'(z)]_i [K(x,z)y]'_i dz  \\[0.3cm]
= \displaystyle\sum_{i=1}^n \int_0^x  [f'(z)]_i y_i dz  \quad \text{($dK(x,z)/d z= diag(1)$ if $z\leq x$, and $=diag(0)$ otherwise)} \\[0.3cm]
= \displaystyle\sum_{i=1}^n  [f(x)]_i y_i dz 
\text{ } = \text{ } \langle f(x), y \rangle_{\mathbb{R}^n}. \smallskip \hfill \blacksquare
\end{array}
\end{align*}

\end{example}

\begin{example}(Function-valued RKHS and its operator-valued kernel) \vspace{1mm} \\ 
\label{exp:fv}
Here we extend Example~\ref{exp:vv} to the case where the output space is infinite dimensional. Let $\mathcal{X}=[0,1]$ and $\mathcal{Y} = L^2(\Omega)$ the space of square integrable functions on  a compact set $\Omega \subset \mathbb{R}$. We denote by $\mathcal{M}$ 
the space of $L^2(\Omega)$-valued functions on $\mathcal{X}$ whose norm 
$ \|g\|_{\mathcal{M}}^2 = \displaystyle\int_{\Omega} \int_\mathcal{X} [g(x)(t)]^2 dx dt$ is finite.

Let $(\mathcal{F};\langle \cdot,\cdot \rangle_{\mathcal{F}})$ be the space of functions from $\mathcal{X}$ to $L^2(\Omega)$ such that:
\begin{equation*}
  \left\{
    \begin{aligned}
    & \mathcal{F} = \big\{f,\ \exists f' = \displaystyle\frac{d f(x)}{d x} \in \mathcal{M}, 
f(x) = \int_0^x f'(z) dz \big\},\\ 
   & \langle f_1, f_2 \rangle_\mathcal{F} = \langle f'_1, f'_2 \rangle_{\mathcal{M}}.
    \end{aligned}
  \right.
\end{equation*}
$\mathcal{F}$ is a function-valued RKHS with the operator-valued kernel $K(x,z) = M_{\varphi(x,z)}$. $M_\varphi$ is the multiplication operator associated 
with the function $\varphi$ where $\varphi(x,z)$ is equal to $x$ if $x \leq z$ and $z$ otherwise. 
Since $\varphi$ is a positive-definite function, 
$K$ is Hermitian and nonnegative. Indeed,
\begin{align*}
\langle K(z,x)^* y , w \rangle_\mathcal{Y} &= \langle  y , K(z,x) w \rangle_\mathcal{Y} = \int_0^1 \varphi(z,x) w(t) y(t) dt = \int_0^1 \varphi(x,z) y(t) z(t) dt \\ & 
= \langle K(x,z) y , w \rangle_\mathcal{Y},
\end{align*}
and 
\begin{align*}
\sum_{i,j} \langle K(x_i,x_j) y_i , y_j \rangle_\mathcal{Y} & = \sum_{i,j}  \int_0^1 \varphi(x_i,x_j) y_i(t) y_j(t) dt \\& 
 = \int_0^1 \sum_{i,j} y_i(t) \varphi(x_i,x_j)  y_j(t) dt \  \geq \ 0 \text{ (since $\varphi \geq 0)$}.
\end{align*}
Now we show that the reproducing property holds for any $f\in \mathcal{F}$, 
$y\in L^2(\Omega)$ and $x\in \mathcal{X}$:
\begin{align*}
\begin{array}{l}
  \langle f, K(x,\cdot)y \rangle_{\mathcal{F}} = \langle f', [K(x,\cdot)y]' \rangle_{\mathcal{M}}\\[0.1cm]
= \displaystyle\int_\Omega \int_0^1  [f'(z)](t) [K(x,z)y]'(t) dz dt \\[0.3cm]
\eqdef \displaystyle\int_\Omega \int_0^x  [f'(z)](t) y(t) dz dt 
=  \displaystyle\int_\Omega  [f(x)](t) y(t) dt \\[0.3cm]
=  \langle f(x), y \rangle_{L^2(\Omega)}. \hfill \blacksquare
\end{array}
\end{align*}
\end{example}

%


Theorem~\ref{th:positiveDefiniteImpliesKernel} states that it is possible 
to construct a pre-Hilbert space of operators from a nonnegative operator-valued 
kernel and with some additional assumptions it can be completed 
to obtain a function-valued reproducing kernel Hilbert space. 
Therefore, it is important to consider the problem of constructing 
nonnegative operator-valued kernels. This is the focus of the next section.


\section{Operator-valued Kernels for Functional Data}
\label{fk}

Reproducing kernels play an important role in statistical learning
theory and functional estimation.  Scalar-valued kernels are widely
used to design nonlinear learning methods which have been successfully
applied in several machine learning
applications~\citep{Scholkopf-2002, Shawetaylor-2004}.  Moreover, their extension to
matrix-valued kernels has helped to bring additional improvements in
learning vector-valued functions~\citep{Micchelli-2005a,
  Reisert-2007, Caponnetto-2006}.  The most common and most successful applications of
matrix-valued kernel methods are in multi-task
learning~\citep{Evgeniou-2005, Micchelli-2005b}, even though some
successful applications also exist in other areas, such as image
colorization~\citep{Minh-2010}, link prediction~\citep{Brouard-2011} 
and network inference~\citep{Lim-2014}. A basic, albeit not
obvious, question which is always present with reproducing kernels
concerns how to build these kernels and what is the optimal kernel
choice. This question has been studied extensively for scalar-valued
kernels, however it has not been investigated
enough in the matrix-valued case.  In the context of multi-task
learning, matrix-valued kernels are constructed from scalar-valued kernels 
which are carried over to the
vector-valued setting by a positive definite matrix
\citep{Micchelli-2005b, Caponnetto-2008}.  

In this section we consider
the problem from a more general point of view.  We are interested in
the construction of operator-valued kernels, generalization of
matrix-valued kernels in infinite dimensional spaces, that perform the
mapping between two spaces of functions and which are suitable for
functional response data. Our motivation is to build operator-valued kernels that are capable of giving rise to nonlinear FDA methods. 
It is worth recalling that previous studies have provided examples of operator-valued kernels with infinite-dimensional output spaces~\citep{Micchelli-2005a, Caponnetto-2008, Carmeli-2010}; however, they did not focus either on building methodological connections with the area of FDA, or on the practical impact of such kernels on real-world applications. 

Motivated by building kernels that capture dependencies between
samples of functional~(infinite-dimensional) response variables, we adopt a FDA modeling formalism.
The design of such kernels will doubtless prove difficult, but it is necessary to develop reliable nonlinear FDA methods. Most FDA methods in the literature are based on linear parametric models. Extending these methods to nonlinear contexts should render them more powerful and efficient. Our line of attack is to construct operator-valued kernels from operators already used to build linear FDA models, particularly those involved in functional response models. Thus, it is important to begin by looking at these models.

\subsection{Linear Functional Response Models}

FDA is an extension of multivariate data analysis suitable when data
are functions.  In this framework, a data is a single function
observation rather than a collection of observations. It is true that
the data measurement process often provides a vector rather than a
function, but the vector is a discretization of a real attribute which
is a function. Hence, a functional datum $i$ is acquired
as a set of discrete measured values, $y_{i1},\ldots,y_{ip}$; the
first task in parametric~(linear) FDA methods is to convert these values
to a function $y_{i}$ with values $y_{i}(t)$ computable for any
desired argument value~$t$. If the discrete values are assumed to be
noiseless, then the process is interpolation; but if they have some
observational error, then the conversion from discrete data to
functions is a regression task (\eg, smoothing)~\citep{Ramsay-2005}.

A functional data model takes the form
$y_{i}=f(x_{i})+\epsilon_{i}$ where one or more of the components
$y_{i}$, $x_{i}$ and $\epsilon_{i}$ are functions.  Three
subcategories of such models can be distinguished: predictors $x_{i}$
are functions and responses $y_{i}$ are scalars; predictors are
scalars and responses are functions; both predictors and responses are
functions. In the latter case, which is the context we face, the function $f$ is a compact operator
between two infinite-dimensional Hilbert spaces.  Most previous works
on this model suppose that the relation between functional responses
and predictors is linear; for more details, see~\citet{Ramsay-2005} and references therein.

For functional input and output data, the functional linear model  commonly found in the literature
is an extension of the multivariate linear one and has the following form:
\begin{equation}
\label{concurrent}
  y(t)=\alpha(t) + \beta(t) x(t) + \epsilon(t),
\end{equation}
where $\alpha$ and $\beta$ are the functional parameters of the model~\citep[Chapter 14]{Ramsay-2005}. This model is known as the
``concurrent model'' where ``concurrent'' means that $y(t)$ only depends on $x$ at $t$. The concurrent model is similar to the varying
coefficient model proposed by~\citet{Hastie-1993} to deal with the
case where the parameter $\beta$ of a multivariate regression model
can vary over time. 
A main limitation of this model is that the response $y$ and the covariate $x$ 
are both functions of the same argument~$t$,~and the influence of a covariate
on the response is concurrent or point-wise in the sense that~$x$ only
influences $y(t)$ through its value $x(t)$ at time $t$. 
To overcome this restriction, 
an extended linear model in which the
influence of a covariate $x$ can involve a range of argument values
$x(s)$ was proposed; it takes the following form:
\begin{equation}
\label{flm}
  y(t)=\alpha(t) + \int x(s)\beta(s,t)ds+\epsilon(t),
\end{equation}
where, in contrast to the concurrent model, the functional parameter $\beta$ is now a function of both $s$ and $t$, and $y(t)$ depends on $x(s)$ for an interval of values of $s$~\citep[Chapter 16]{Ramsay-2005}.  Estimation of the parameter function $\beta(\cdot,\cdot)$ is an inverse problem and requires regularization. Regularization can be implemented in a variety of ways, for
example by penalized splines~\citep{James-2002} or by truncation of series expansions~\citep{Mueller-2005funcional}. A review of functional response models can be found in~\citet{Chiou-2004}.

The operators involved in the functional data models described above are the multiplication operator~(Equation~\ref{concurrent}) and the integral operator~(Equation~\ref{flm}). We think that operator-valued kernels constructed using these operators could be a valid alternative to extend linear FDA methods to nonlinear settings. In Subsection~\ref{ex} we provide examples of multiplication and integral operator-valued kernels. Before that, we identify building schemes that can be common to many operator-valued kernels and applied to functional data.%

\subsection{Operator-valued Kernel Building Schemes}

\begin{figure}
\centering
\begin{tabular}{c|c}
 Function combination & Operator combination \\[5pt]
\hline
\\[-5pt]
\xymatrix{\\ **[l] \mathcal{X} \times \mathcal{X} \ar@{->}[r]_-{z(x_1,x_2)\ } & 
\mathcal{Z} \ar@{->}[r]_-{T(z)}  & \mathcal{L(Y)} } &

\xymatrix{  \mathcal{X} \ar@{->}[r]_-{T(x_1)} & \mathcal{L(X,Y)} \ar@{->}[d] \\ 
 &  \mathcal{L(X,Y)} &  **[l] \times \ \ \mathcal{L(X,Y)}^* \ar@{->}[r]_-{T(x_1)T(x_2)^*} &  \mathcal{L(Y)} \\
  \mathcal{X} \ar@{->}[rr]_-{T(x_2)} &  & **[l] \mathcal{L(X,Y)} \ar@<3.5ex>@{->}[u]  & 
}
\end{tabular}
   \caption{Illustration of building an operator-valued kernel from $\x \times \x$ to $\mathcal{L(Y)}$ using
     a combination of functions or a combination of operators. (left) The operator-valued kernel is constructed by combining two functions ($x_1$ and $x_2$) and by applying a positive $\mathcal{L(Y)}$-valued mapping $T$ to the combination. (right) the operator-valued kernel is generated by combining two operators ($T(x_1)$ and $T(x_2)^*$)  built from an $\mathcal{L(X,Y)}$-valued mapping $T$.}
  \label{figkernel}
\end{figure}
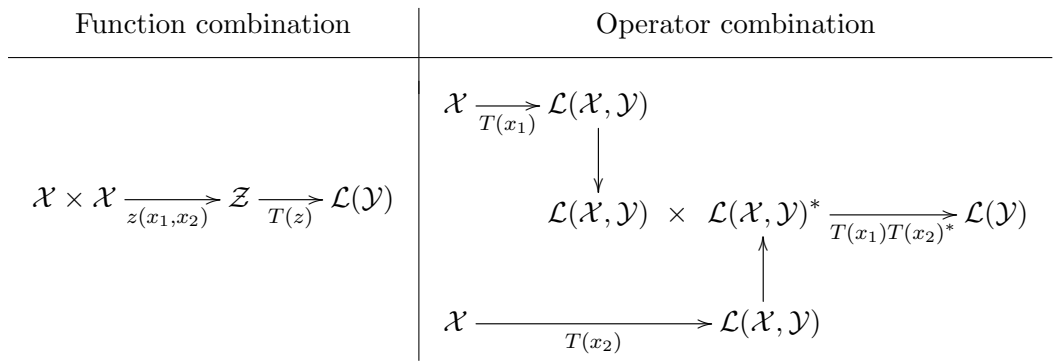

In our context, constructing an operator-valued kernel turns out to build an operator that maps a couple of functions to a function:  in $\mathcal{X}\times\mathcal{X}\rightarrow\mathcal{L(Y)}$ from two functions $x_1$ and $x_2$ in $\mathcal{X}$.  This  can be performed in one of two
ways: either combining the two functions $x_{1}$ and $x_{2}$ into a variable $z \in \mathcal{Z}$ and then adding an operator function $T: \mathcal{Z} \longrightarrow \mathcal{L(Y)}$ 
that performs the mapping from space $\mathcal{Z}$ to $\mathcal{L(Y)}$, or building an $\mathcal{L(X,Y)}$-valued function $T$, where $\mathcal{L(X,Y)}$ is the set 
of bounded operators from $\mathcal{X}$~to~$\mathcal{Y}$, and then combining the resulting operators $T(x_1)$ and $T(x_2)$ to obtain the operator in $\mathcal{L(Y)}$. In the latter case, a natural way to combine $T(x_1)$ and $T(x_2)$ is to use the composition operation and the kernel $K(x_1,x_2)$ will be equal to $T(x_1)T(x_2)^*$. 
Figure~\ref{figkernel} describes the construction of an operator-valued kernel function using 
the two schemes which are based on combining functions~($x_1$ and $x_2$) or operators~($T(x_1)$ and $T(x_2)$), respectively.
Note that separable operator-valued kernels~\citep{Alvarez-2012}, which are kernels that 
can be formulated as a product of a scalar-valued kernel function for the input space alone 
and an operator that encodes the interactions between the outputs, are a particular case of 
the function combination building scheme, when we take $\mathcal{Z}$ as the set of real numbers 
$\mathbb{R}$ and the scalar-valued kernel as combination function.
In contrast, the operator combination scheme is particularly amenable to the design of nonseparable operator-valued kernels. This scheme was already used in various problems of operator theory, system theory and interpolation~\citep{Alpay-1997, Dym-1989}.
%

To build an operator-valued kernel and then construct a
function-valued reproducing kernel Hilbert space, the operator $T$
is of crucial importance. Choosing $T$ presents two major
difficulties. Computing the adjoint operator is not always easy to do,
and then, not all operators verify the Hermitian condition of the
kernel. On the other hand, since the kernel must be nonnegative, we suggest
to construct operator-valued kernels from positive definite
scalar-valued kernels which can be the reproducing kernels of real-valued
Hilbert spaces. In this case, the reproducing property of the
operator-valued kernel allows us to compute an inner product in a space
of operators by an inner product in a space of functions which can
be, in turn, computed using the scalar-valued kernel. The operator-valued kernel allows
the mapping between a space of functions and a space of operators,
while the scalar one establishes the link between the space of
functions and the space of measured values. 
It is also useful to define combinations of nonnegative operator-valued kernels that
allow to build a new nonnegative one.

\subsection{Combinations of Operator-valued Kernels}
\label{subsec:comb}
We have shown in Section~\ref{sec:frkhs} that there is a bijection
between nonnegative operator-valued kernels and function-valued
reproducing kernel Hilbert spaces. So, as in the scalar case, it will
be helpful to characterize algebraic transformations, like sum and
product, that preserve the nonnegativity of operator-valued
kernels. Theorem~\ref{combinaison noyaux} stated below gives some
building rules to obtain a positive operator-valued kernel from
combinations of positive existing ones. %
Similar results for the case of matrix-valued kernels can be
found in \citet{Reisert-2007}, and for a more general
context we refer the reader to~\citet{Caponnetto-2008} and~\citet{Carmeli-2010}.
In our setting, assuming $H$ and $G$ be two nonnegative kernels constructed as
described in the previous subsection, we are interested in constructing a nonnegative 
kernel $K$ from $H$ and $G$. 
\begin{theo}
 \label{combinaison noyaux}
 \label{th:constructionOfKernels}
Let  $H:\mathcal{X} \times \mathcal{X}
  \longrightarrow \mathcal{L(Y)}$ and 
$G:\mathcal{X} \times \mathcal{X}
  \longrightarrow \mathcal{L(Y)}$ two  nonnegative operator-valued kernels
%
 \begin{enumerate}[\normalfont (i)]
   \item \label{enum2:i} $K \equiv H+G$ is a nonnegative kernel, 
     %
   \item \label{enum2:ii} if $H(w,z)G(w,z) = G(w,z)H(w,z)$, $\forall w,z \in \mathcal{X}$, then $K \equiv HG$ is a nonnegative kernel,
     %
   \item \label{enum2:iii} 
   $K \equiv T H T^*$ 
	is a nonnegative kernel for any $\mathcal{L(Y)}$-valued function $T(\cdot)$.
   \end{enumerate}
\end{theo}
\textbf{Proof}: Obviously (i) follows from the linearity of the inner product. (ii) can be proved by showing that  
the ``element-wise'' multiplication of two positive block operator matrices can be positive~(see below). 
For the proof of (iii), we observe that
\begin{equation*}
 K(w,z)^* = [T(z)H(w,z)T(w)^*]^* = T(w)H(z,w)T(z)^* = K(z,w),
\end{equation*}
and
\vspace{-0.3cm}
\begin{align*}
 \sum_{i,j}\langle K(w_{i},w_{j})u_{i},u_{j}\rangle  & = 
\sum_{i,j}\langle T(w_{j}) H(w_{i},w_{j}) T(w_{i})^*u_{i},u_{j}\rangle \\ & =
\sum_{i,j}\langle H(w_{i},w_{j})T(w_{i})^*u_{i},T(w_{j})^*u_{j}\rangle,
\end{align*}
which implies the nonnegativity of the kernel $K$ since $H$ is nonnegative.

\vspace{1mm}
%
To prove (ii), i.e., the kernel $K\equiv HG$ is nonnegative in the case where $H$ and $G$ are nonnegative kernels such that $H(w,z)G(w,z) = G(w,z)H(w,z)$, $\forall w,z \in \mathcal{X}$, we show below that the block operator matrix $\mathbf{K}$ associated to the operator-valued kernel $K$ for a given set $\{w_i\}, i=1,\ldots,n$ with $n\in\mathbb{N}$, is positive. By construction, we have $\mathbf{K} = \mathbf{H} \circ \mathbf{G}$ where $\mathbf{H}$ and $\mathbf{G}$ are the block operator kernel matrices corresponding to the kernels $H$ and $G$, and `$\circ$' denotes the ``element-wise'' multiplication defined by $(\mathbf{H}\circ \mathbf{G})_{ij} = H(w_i,w_j)G(w_i,w_j)$. $\mathbf{K}$, $\mathbf{H}$ and $\mathbf{G}$ are all in $\in \mathcal{L}(\mathcal{Y}^n)$.

Since the kernels $H$ and $G$ are Hermitian and $HG = GH$, it is easy to see that
\begin{align*}
(\mathbf{K}^*)_{ij} &= (\mathbf{K}_{ji})^* = K(w_j,w_i)^* = \big(H(w_j,w_i)G(w_j,w_i)\big)^* = G(w_j,w_i)^*H(w_j,w_i)^* \\
&= G(w_i,w_j)H(w_i,w_j) = H(w_i,w_j)G(w_i,w_j) \\
& = \mathbf{K}_{ij}.
\end{align*}
Thus, $\mathbf{K}$ is self-adjoint. It remains, then, to prove that $\langle \mathbf{K} \mathbf{u}, \mathbf{u} \rangle \geq 0$, $\forall \mathbf{u} \in \mathcal{Y}^n$, in order to show the positivity of $\mathbf{K}$.

The ``element-wise'' multiplication can be rewritten as a tensor product. Indeed, we have
\begin{align*}
\mathbf{K} = \mathbf{H} \circ \mathbf{G} = \mathbf{L}^* (\mathbf{H} \otimes \mathbf{G}) \mathbf{L},
\end{align*}
where $\mathbf{L}: \mathcal{Y}^n \longrightarrow \mathcal{Y}^n \otimes \mathcal{Y}^n$ is the mapping defined by $\mathbf{L} \mathbf{e_i} = \mathbf{e_i} \otimes \mathbf{e_i}$ for an orthonormal basis $\{\mathbf{e_i}\}$ of the separable Hilbert space $\mathcal{Y}^n$, and $\mathbf{H}\otimes \mathbf{G}$ is the tensor product defined by $(\mathbf{H}\otimes \mathbf{G}) (\mathbf{u}\otimes\mathbf{v}) = \mathbf{H}\mathbf{u}\otimes\mathbf{G}\mathbf{v}$, $\forall \mathbf{u},\mathbf{v} \in \mathcal{Y}^n$.
To see this, note that
\begin{align*}
\langle \mathbf{L}^* (\mathbf{H} \otimes \mathbf{G}) \mathbf{L} \mathbf{e_i}, \mathbf{e_j} \rangle 
&= \langle (\mathbf{H} \otimes \mathbf{G}) \mathbf{L} \mathbf{e_i}, \mathbf{L} \mathbf{e_j} \rangle 
=  \langle (\mathbf{H} \otimes \mathbf{G}) (\mathbf{e_i} \otimes\mathbf{e_i}), \mathbf{e_j} \otimes \mathbf{e_j} \rangle \\
&= \langle \mathbf{H} \mathbf{e_i} \otimes \mathbf{G}\mathbf{e_i}, \mathbf{e_j} \otimes \mathbf{e_j} \rangle
 =  \langle \mathbf{H} \mathbf{e_i}, \mathbf{e_j} \rangle \langle \mathbf{G} \mathbf{e_i}, \mathbf{e_j} \rangle\\
 &=  \mathbf{H}_{ij} \mathbf{G}_{ij} = \langle  (\mathbf{H} \circ \mathbf{G}) \mathbf{e_i}, \mathbf{e_j} \rangle.
\end{align*}
Now since $H$ and $G$ are positive, we have 
\begin{align*}
\langle \mathbf{K} \mathbf{u}, \mathbf{u} \rangle 
&= \langle  \mathbf{L}^* (\mathbf{H} \otimes \mathbf{G}) \mathbf{L} \mathbf{u}, \mathbf{u} \rangle
=  \langle  \mathbf{L}^* (\mathbf{H}^{\frac{1}{2}} \mathbf{H}^{\frac{1}{2}}  \otimes \mathbf{G}^{\frac{1}{2}}  \mathbf{G}^{\frac{1}{2}} ) \mathbf{L} \mathbf{u}, \mathbf{u} \rangle \\
&= \langle  \mathbf{L}^* (\mathbf{H}^{\frac{1}{2}}  \otimes \mathbf{G}^{\frac{1}{2}}) (\mathbf{H}^{\frac{1}{2}} \otimes \mathbf{G}^{\frac{1}{2}} ) \mathbf{L} \mathbf{u}, \mathbf{u} \rangle
= \langle (\mathbf{H}^{\frac{1}{2}} \otimes \mathbf{G}^{\frac{1}{2}} ) \mathbf{L} \mathbf{u},   (\mathbf{H}^{\frac{1}{2}}  \otimes \mathbf{G}^{\frac{1}{2}})^* \mathbf{L} \mathbf{u} \rangle \\
& = \langle (\mathbf{H}^{\frac{1}{2}} \otimes \mathbf{G}^{\frac{1}{2}} ) \mathbf{L} \mathbf{u},   (\mathbf{H}^{\frac{1}{2}}  \otimes \mathbf{G}^{\frac{1}{2}}) \mathbf{L} \mathbf{u} \rangle
= \|(\mathbf{H}^{\frac{1}{2}}  \otimes \mathbf{G}^{\frac{1}{2}}) \mathbf{L} \mathbf{u}\|^2 \geq 0.
\end{align*}
This concludes the proof.

\hfill $\blacksquare$ 

\subsection{Examples of Nonnegative Operator-valued Kernels}
\label{ex}

We provide here examples of operator-valued kernels for
functional response data. All these examples deal with operator-valued
kernels constructed following the schemes described above and assuming
that $\mathcal{Y}$ is an infinite-dimensional function space.
Motivated by building kernels suitable for functional data, the first
two examples deal with operator-valued kernels constructed from the
multiplication and the integral self-adjoint operators
in the case where $\mathcal{Y}$ is the Hilbert space
$L^2(\Omega_{y})$ of square integrable functions on $\Omega_{y}$
endowed with the inner product $\langle \phi,\psi\rangle=
\int_{\Omega_y} \phi(t)\psi(t)dt$.  
We think that these kernels represent an interesting alternative to extend linear functional models to
nonlinear settings. The third example based on the composition operator
shows how to build such kernels from non self-adjoint operators 
(this may be relevant when the functional linear model is based on a 
non self-adjoint operator). 
It also 
illustrates the kernel combination defined in Theorem~\ref{combinaison noyaux}(iii). 
\vspace{-0.1cm}
\begin{enumerate}
 \item Multiplication operator:

   In~\citet{Kadri-2010}, the authors attempted to extend the widely
   used Gaussian kernel to functional data domain using a
   multiplication operator and assuming that input and output data
   belong to the same space of functions. 
   Here we consider a slightly different setting, where the input space $\mathcal{X}$ can be 
   different from the output space $\mathcal{Y}$.

   A multiplication operator on $\mathcal{Y}$  is defined as follows:
   \begin{equation}
     \nonumber
     \begin{array}{ccll} T^{h}: & \mathcal{Y} & \longrightarrow &
       \mathcal{Y}\\
       & y & \longmapsto & T^{h}_{y}\ \ ;\ \ T^{h}_{y}(t) \triangleq h(t)y(t).
     \end{array}
   \end{equation}
   The operator-valued kernel $K(\cdot,\cdot)$ is the following:
   \begin{eqnarray*}
     \begin{array}{ccll} K:& \mathcal{X} \times
       \mathcal{X} & \longrightarrow &
       \mathcal{L(Y)}\\
       & x_{1} , x_{2} & \longmapsto & k_x(x_1,x_2)T^{ k_y},
     \end{array}
   \end{eqnarray*}
   where $k_x(\cdot,\cdot)$ is a positive definite scalar-valued kernel and $k_y$ a positive real 
function. It is easy to see that $\langle T^{h}x,y\rangle = \langle x,T^{h}y\rangle $, then
   $T^{h}$ is a self-adjoint operator. Thus $K(x_{2} ,
   x_{1})^{*} = K(x_{2} , x_{1})$ and $K$
   is Hermitian since $K(x_{1} , x_{2}) =
   K(x_{2} , x_{1})$.

   Moreover, we have 
  \begin{align*}
       \displaystyle\sum\limits_{i,j}&\langle K(x_{i},x_{j})y_{i},y_{j}\rangle_{\mathcal{Y}} 
        =  \displaystyle\sum\limits_{i,j } k_x(x_i,x_j) \langle k_y(\cdot)y_{i}(\cdot),y_{j}(\cdot)\rangle_{\mathcal{Y}} 
       \\  &= \displaystyle\sum\limits_{i,j} k_x(x_i,x_j) \int k_y(t)y_{i}(t)y_{j}(t)dt 
    = \displaystyle \int \sum\limits_{i,j} y_{i}(t)[k_x(x_i,x_j) k_y(t)]y_{j}(t)dt \geq 0,
   \end{align*}
   since the product of two positive-definite scalar-valued kernels is also positive-definite. Therefore
   $K$ is a nonnegative operator-valued kernel.\\[-0.2cm]

\item Hilbert-Schmidt integral operator:

A Hilbert-Schmidt integral operator on $\mathcal{Y}$  associated with a kernel $h(\cdot,\cdot)$ 
is defined as follows:
\begin{equation}
  \nonumber
  \begin{array}{ccll} T^{h}: & \mathcal{Y} & \longrightarrow &
    \mathcal{Y}\\
    & y & \longmapsto & T^{h}_{y}\ \ ;\ \ T^{h}_{y}(t) \triangleq \int h(s,t)y(s) ds.
  \end{array}
\end{equation}
In this case, an operator-valued kernel $K$ is a
Hilbert-Schmidt integral operator associated with positive definite scalar-valued 
kernels $k_x$ and $k_y$, and it takes the following form:
\begin{eqnarray*}
  \begin{array}{ccll} K(x_{1},x_{2})[\cdot]:& \mathcal{Y} & \longrightarrow &
    \mathcal{Y}\\
    & f & \longmapsto & g
  \end{array}
\end{eqnarray*}
where $g(t) = k_x(x_1,x_2)\displaystyle\int k_y(s,t)f(s)ds$.

The Hilbert-Schmidt integral operator is self-adjoint if $k_y$ is Hermitian. 
This condition is verified 
and then it is easy to check that $K$ is also Hermitian.
$K$ is nonnegative since 
 \begin{equation*}
\sum\limits_{i,j}\langle K(x_{i},x_{j})y_{i},y_{j}\rangle_{\mathcal{Y}} 
 =   \iint\sum\limits_{i,j} y_{i}(s)[k_x(x_i,x_j) k_y(s,t)]y_{j}(t) ds dt,
\end{equation*}
which is positive because of the positive-definiteness of the scalar-valued kernels $k_x$ and $k_y$.\\[-0.2cm]

\item Composition operator:

Let $\varphi$ be an analytic map. The composition operator associated with $\varphi$ is 
the linear map:
\begin{equation*}
C_{\varphi}: f \longmapsto f \circ \varphi
\end{equation*}
First, we look for an expression of the adjoint of the composition operator $C_{\varphi}$ acting on $\mathcal{Y}$
in the case where $\mathcal{Y}$ is a scalar-valued RKHS of functions on $\Omega_{y}$ and $\varphi$ an 
analytic map of $\Omega_{y}$ into itself. For any $f$ in the space $\mathcal{Y}$ associated 
with the real kernel $k$,
\begin{equation*}
\begin{array}{ccccc}
 \langle f,C_{\varphi}^{*}k_{t}(\cdot)\rangle &=& \langle C_{\varphi}f,k_{t}\rangle &=& 
 \langle f \circ \varphi, k_{t} \rangle \\ &=& f(\varphi(t)) &=& 
 \langle f, k_{\varphi(t)} \rangle.
\end{array}
\end{equation*}
This is true for any $f \in \mathcal{Y}$ and then $C_{\varphi}^{*}k_{t} = k_{\varphi(t)}$. In a similar 
way, $C_{\varphi}^{*}f$ can be computed at each point of the function $f$:
\begin{equation*}
(C_{\varphi}^{*}f)(t) = \langle C_{\varphi}^{*}f, k_{t} \rangle = 
\langle f, C_{\varphi} k_{t} \rangle = \langle f, k_{t}\circ \varphi \rangle
\end{equation*}
Once we have expressed the adjoint of a composition operator in a reproducing kernel Hilbert space, 
we consider the following operator-valued kernel:
\begin{eqnarray*}
  \begin{array}{ccll} K:& \mathcal{X} \times
    \mathcal{X} & \longrightarrow &
    \mathcal{L(Y)}\\
    & x_{1} , x_{2} & \longmapsto & C_{\psi(x_1)} C_{\psi(x_2)}^{*}
  \end{array}
\end{eqnarray*}

where $\psi(x_{1})$ and $\psi(x_{2})$ are maps of $\Omega_{y}$ into itself. It is easy to see
that the kernel $K$ is Hermitian. Using Theorem
\ref{th:constructionOfKernels}(iii) we obtain the nonnegativity
property of the kernel.

\end{enumerate}

\subsection{Multiple Functional Data and Kernel Feature Map}

Until now, we discussed operator-valued kernels and their
corresponding RKHS from the perspective of extending~\citet{Aronszajn-1950} 
pioneering work from scalar-valued or vector-valued cases to the function-valued case.  However, it is also
interesting to explore these kernels from a feature space point of
view~\citep{Sch-1999, Caponnetto-2008}. In this subsection, we provide
some ideas targeted at advancing the understanding of feature spaces
associated with operator-valued kernels and we show how these kernels
can design more suitable feature maps than those associated with
scalar-valued kernels, especially when input data are 
infinite dimensional objects like curves.
To explore the potential of adopting an operator-valued kernel feature
space approach, we consider a supervised learning problem with
multiple functional data where each observation is composed of more
than one functional variable~\citep{Kadri-2011b,Kadri-2011c}. 
Working with multiple functions allows to deal in a natural way with a
lot of applications. There are many practical situations where a number of
potential functional covariates are available to explain a response variable. 
For example, in audio and speech processing where signals are converted
into different functional features providing information about their
temporal, spectral and cepstral characteristics, 
or in meteorology where the interaction effects between various 
continuous variables (such as temperature, precipitation, and winds) is of particular interest.

Similar to the scalar case, operator-valued kernels provide an elegant way of 
dealing with nonlinear algorithms by reducing them to linear ones in some feature 
space $F$ nonlinearly related to input space. 
A feature map associated with 
an operator-valued kernel $K$ is a continuous function 
\begin{equation*}
 \Phi: \mathcal{X} \times \mathcal{Y} \longrightarrow \mathcal{L(X,Y)},
\end{equation*}
such that for every $x_1, x_2 \in \mathcal{X}$ and $y_1, y_2 \in \mathcal{Y}$
\begin{equation*}
 \langle K(x_1,x_2)y_1, y_2 \rangle_{\mathcal{Y}} = \langle \Phi(x_1,y_1), \Phi(x_2,y_2) \rangle_{\mathcal{L(X,Y)}},
\end{equation*}
where $\mathcal{L(X,Y)}$ is the set of linear mappings from $\mathcal{X}$ into $\mathcal{Y}$. 
By virtue of this property, $\Phi$ is called a \textit{feature map associated with} $K$. 
Furthermore, from the reproducing property, it follows that in particular
\begin{equation*}
\langle K(x_1,\cdot)y_1, K(x_2,\cdot)y_2 \rangle_{\mathcal{F}} = \langle K(x_1,x_2)y_1, y_2 \rangle_{\mathcal{Y}},
\end{equation*}
which means that any operator-valued kernel admits a feature map 
representation $\Phi$ with a feature space $\mathcal{F} \subset \mathcal{L(X,Y)}$ 
defined by $\Phi(x_1,y_1)= K(x_1,\cdot)y_1$, and corresponds to an inner product in another space. 

From this feature map perspective, we study the geometry of a feature
space associated with an operator-valued kernel and we compare it with
the geometry obtained by a scalar-valued kernel.  More precisely, we
consider two reproducing kernel Hilbert spaces $\mathcal{F}$
and $\mathcal{H}$.  $\mathcal{F}$ is a RKHS of function-valued
functions on $\mathcal{X}$ with values in $\mathcal{Y}$.
$\mathcal{X} \subset (L^2(\Omega_x))^p$ \footnote{$p$ is the number of functions that represent input data. In the field of FDA, such data are called multivariate functional data.}, $\mathcal{Y} \subset
L^2(\Omega_y)$ and let $K$ be the reproducing operator-valued kernel
of $\mathcal{F}$.  $\mathcal{H}$ is also a RKHS, but of scalar-valued
functions on $\mathcal{X}$ with values in $\mathbb{R}$, and $k$ its reproducing
scalar-valued kernel. The mappings $\Phi_K$ and $\Phi_k$ associated,
respectively, with the kernels $K$ and $k$ are defined as follows
\begin{equation}
\nonumber
  \Phi_K: (L^2)^p \rightarrow \mathcal{L}((L^2)^p,L^2),\quad x \mapsto K(x,\cdot)y,
\end{equation}
and
\begin{equation}
\nonumber
 \Phi_k: (L^2)^p \rightarrow \mathcal{L}((L^2)^p,\mathbb{R}),\quad x \mapsto k(x,\cdot).
\end{equation}
These feature maps can be seen as a mapping of the input data $x_i$,
which are vectors of functions in $(L^2)^p$ , into a feature space in
which the inner product can be computed using the kernel functions. This
idea leads to design nonlinear methods based on linear ones in the
feature space. In a supervised classification problem for example,
since kernels map input data into a higher dimensional space,
kernel methods deal with this problem by finding a linear separation
in the feature space.
We now compare the dimension of feature spaces obtained by the maps
$\Phi_K$ and $\Phi_k$. To do this, we adopt a functional data analysis
point of view where observations are composed of sets of functions.
Direct understanding of this FDA viewpoint comes from the
consideration of the ``atom'' of a statistical analysis.  In a basic
course in statistics, atoms are ``numbers'', while in multivariate
data analysis the atoms are vectors and methods for understanding
populations of vectors are the focus. FDA can be viewed as the
generalization of this, where the atoms are more complicated objects,
such as curves, images or shapes represented by
functions~\citep{Zhao-2004}.  Based on this, the dimension of the
input space is $p$ since $x_i\in (L^2)^p$ is a vector of $p$
functions. The feature space obtained by the map $\Phi_k$ is a space
of functions, so its dimension from a FDA viewpoint is equal to one. 
The map $\Phi_K$ projects the
input data into a space of operators $\mathcal{L(X,Y)}$. This means
that using the operator-valued kernel $K$ corresponds to mapping the
functional data $x_i$ into a higher, possibly infinite, dimensional
space $(L^2)^d$ with $d\rightarrow \infty$.  In a binary functional
classification problem, we have higher probability to achieve linear
separation between the classes by projecting the functional data into
a higher dimensional feature space rather than into a lower one (Cover's 
theorem), that is why we think that it is more suitable to use
operator-valued than scalar-valued kernels in this context.


\section{Function-valued Function Learning}
\label{fvfe}

In this section, we consider the problem of estimating an
unknown function $F$ such that $F(x_i)=y_i$ when observed data 
$(x_{i}(s),y_{i}(t))_{i=1}^{n} \in \mathcal{X} \times
\mathcal{Y}$ are assumed to be elements of the space of square integrable functions $L^2$. 
$X=\{x_{1},\ldots,x_{n}\}$ denotes the training set with
corresponding targets $Y=\{y_{1},\ldots,y_{n}\}$. Since
$\mathcal{X}$ and $\mathcal{Y}$ are spaces of functions, the
problem can be thought of as an operator estimation problem, where the
desired operator maps a Hilbert space of factors to a Hilbert space of
targets. Among all functions in a linear space of operators
$\mathcal{F}$, an estimate $\widetilde{F} \in \mathcal{F}$ of
$F$ may be obtained by minimizing:
\begin{equation}
\nonumber
 \widetilde{F}  =  \arg\min\limits_{F \in \mathcal{F}} \sum\limits_{i=1}^{n}\|y_{i}-F(x_{i})\|_{\mathcal{Y}}^{2}.
\end{equation}
Depending on $\mathcal{F}$, this problem can be ill-posed and a
classical way to turn it into a well-posed problem is to use a
regularization term. Therefore, we may consider the solution of the
problem as the function $\widetilde{F}  \in \mathcal{F}$ that minimizes:
\begin{equation}
\label{mp}
 \widetilde{F} _\lambda =  \arg\min\limits_{F \in \mathcal{F}} \sum\limits_{i=1}^{n}\|y_{i}-F(x_{i})\|_{\mathcal{Y}}^{2}
     +\lambda\|F\|_{\mathcal{F}}^{2},
\end{equation}
where $\lambda \in \mathbb{R}^+$ is a regularization parameter.
Existence of $\widetilde{F} _\lambda$ in the optimization problem~(\ref{mp}) is guaranteed for $\lambda>0$ by the generalized Weierstrass theorem and one of its corollary that
we  recall from~\citet{Kurdila-2005}.
\begin{theo}
 Let $\mathcal{Z}$ be a reflexive Banach space and $\mathcal{C} \subseteq \mathcal{Z}$ a weakly closed
 and bounded set. Suppose $J: \mathcal{C} \to \mathbb{R}$ is a proper lower
semi-continuous function. Then $J$ is bounded from below and has a minimizer
on $\mathcal{C}$. 
\end{theo}
\begin{corrol}
Let $\mathcal{H}$ be a Hilbert space and $J : \mathcal{H} \to \mathbb{R}$ is a strongly lower semi-continuous, convex and coercive function. Then $J$ is bounded from below and attains a minimizer.
\end{corrol}
This corollary can be straightforwardly applied to problem~(\ref{mp})  by defining: 
$$J_\lambda(F) = \sum\limits_{i=1}^{n}\|y_{i}-F(x_{i})\|_{\mathcal{Y}}^{2}
+\lambda\|F\|_{\mathcal{F}}^{2},$$
where $F $ belongs to the Hilbert space $\mathcal{F}$. It is easy to note
that $J_\lambda$ is continuous and convex. Besides, $J_\lambda$ is coercive for
$\lambda >0$ since $\|F\|^2_{\mathcal{F}}$ is coercive and the sum involves
only positive terms. Hence $\widetilde{F}_\lambda =  \displaystyle\arg\min_{F \in \mathcal{F}} J_\lambda(F)$ exists.

\subsection{Learning Algorithm} 

We are now interested in solving the minimization
problem~(\ref{mp})  in a reproducing kernel Hilbert space $\mathcal{F}$ of
function-valued functions. In the scalar case, it is well-known
that under general conditions on real-valued RKHS, the solution of this
minimization problem can be written as:
\begin{equation}
  \nonumber \widetilde{F} (x)=\sum\limits_{i=1}^{n}\alpha_{i}k(x_i,x), 
\end{equation}
where $\alpha_i \in \mathbb{R}$ and $k$ is the reproducing kernel of a real-valued Hilbert space~\citep{Wahba-1990}. An
extension of this solution to the domain of functional data analysis takes
the following form:
\begin{equation}
\label{repth}
    \widetilde{F} (\cdot)=\sum\limits_{i=1}^{n}K(x_{i},\cdot)u_{i},
  \end{equation}
where $u_i(\cdot)$ are in $\mathcal{Y}$ and the reproducing kernel 
 $K$ is a nonnegative operator-valued function.
With regards to the classical representer theorem, 
here the kernel $K$ outputs an operator and the ``weights'' $u_i$ are functions. 
A proof of the representer theorem in the case of function-valued reproducing kernel Hilbert spaces is given in Appendix~\ref{App:RepresenterTheorem}~(see also~\citealp{Micchelli-2005a}). 

Substituting~(\ref{repth}) in~(\ref{mp}) and using the reproducing property of $\mathcal{F}$, 
we come up with the following minimization problem over the scalar-valued functions 
$u_i \in \mathcal{Y}$ 
($\mathbf{u}$ is the vector of functions $(u_i)_{i=1,\dots,n} \in (\mathcal{Y})^n$) 
rather than the function-valued function (or operator) $F$: 
\begin{equation}
  \label{mp1}
  \mathbf{\widetilde{u}}_\lambda=\arg\min\limits_{\mathbf{u}\in(\mathcal{Y})^n}
   \sum\limits_{i=1}^{n}\|y_{i}-\sum\limits_{j=1}^{n}
  K(x_{i},x_{j})u_{j}\|_{\mathcal{Y}}^{2}  
   +\lambda
  \sum\limits_{i,j}^{n}\langle K(x_{i},x_{j})u_{i},u_{j}\rangle_{\mathcal{Y}} .
\end{equation}
%
Problem~(\ref{mp1}) can be solved in three ways:
\begin{enumerate}
  \item Assuming that the observations are made on a regular grid
    $\{t_1,\ldots,t_m\}$, one can first discretize the functions $x_i$
    and $y_i$ and then solve the problem using multivariate data
    analysis techniques~\citep{Kadri-2010}. However, as this is
    well-known in the FDA domain, this has the drawback of not taking
    into consideration the relationships that exist between samples.
  \item The second way consists in considering the output space
    $\mathcal{Y}$ to be a scalar-valued reproducing Hilbert
    space. In this case, the functions $u_i$ can be approximated
    by a linear combination of a scalar-valued kernel $\hat{u}_i =
    \sum_{l=1}^m\alpha_{il}k(s_l,\cdot) $ and then the problem~(\ref{mp1}) 
becomes a minimization problem over the real values $\alpha_{il}$
    rather than the discrete values
    $u_i(t_1),\ldots,u_i(t_m)$. In the FDA literature, a
    similar idea has been adopted by~\citet{Ramsay-2005} and
    by~\citet{Prchal-2007} who expressed not only the functional
    parameters $u_i$ but also the observed input and output data
    in a basis functions specified a priori (\eg{}, Fourier basis or
    B-spline basis).
  \item Another possible way to solve the minimization
    problem~(\ref{mp1}) is to compute its derivative using the
    directional derivative and setting the result to zero to find an
    analytic solution of the problem. It follows that the vector of
    functions $\mathbf{u} \in \mathcal{Y}^n$ satisfies the system of
    linear operator equations:
    \begin{equation}
      \label{sloe}
      (\mathbf{K}+\lambda I)\mathbf{u} = \mathbf{y},
    \end{equation}
    where $\mathbf{K} = [K(x_i,x_j)]_{i,j=1}^n$ is a
    $n\times n$ block operator kernel matrix ($\mathbf{K}_{ij} \in
    \mathcal{L(Y)}$) and $\mathbf{y} \in \mathcal{Y}^n$ the vector of
    functions $(y_i)_{i=1}^n$. In this work, we are interested in this
    third approach which extends to functional data analysis domain
    results and properties known from multivariate statistical
    analysis.  One main obstacle for this extension is the inversion
    of the block operator kernel matrix $\mathbf{K}$. Block operator
    matrices generalize block matrices to the case where the block
    entries are linear operators between infinite dimensional Hilbert
    spaces.  These matrices and their inverses arise in some areas of
    mathematics~\citep{Tretter-2008} and signal
    processing~\citep{Asif-2005}.  In contrast to the multivariate
    case, inverting such matrices is not always feasible in infinite
    dimensional spaces.  To overcome this problem, we study the
    eigenvalue decomposition of a class of block operator kernel
    matrices obtained from operator-valued kernels having the
    following form:
    \begin{equation}
      \label{kern}
      K(x_i,x_j) = g(x_i,x_j)T,\quad \forall x_i,x_j \in \mathcal{X},
    \end{equation}
    where $g$ is a scalar-valued kernel and $T$ is an operator in
    $\mathcal{L(Y)}$.  This separable kernel construction is adapted
    from~\citet{Micchelli-2005a, Micchelli-2005b}. The choice of $T$
    depends on the context. For multi-task kernels, $T$ is a finite
    dimensional matrix which models relations between tasks. In
    FDA,~\citet{Lian-2007} suggested the use of the identity operator,
    while in~\citet{Kadri-2010} the authors showed that it is better
    to choose other operators than identity to take into
    account functional properties of the input and output spaces. They
    introduced a functional kernel based on the multiplication
    operator. In this work, we are more interested in kernels
    constructed from the integral operator. This seems to be a
    reasonable choice since functional linear model~(see
    Equation~\ref{flm}) are based on this
    operator~\citep[Chapter~16]{Ramsay-2005}.  So we can consider for example the
    following positive definite operator-valued kernel:
    \begin{equation}
      \label{kernn}
      (K(x_i,x_j)y)(t) = g(x_i,x_j)\int_{\Omega_y} e^{-|t-s|} y(s) ds,
    \end{equation}
    where $y\in \mathcal{Y} = L^2(\Omega_{y})$ and $\{s,t\}\in \Omega_y =[0,1]$.
    Note that a similar kernel was proposed as an example in~\cite{Caponnetto-2008}
    for linear spaces of functions from $\mathbb{R}$ to
    $\mathcal{G}_{y}$.

The $n\times n$ block operator kernel matrix $\mathbf{K}$ of
 operator-valued kernels having the form~(\ref{kern}) can be expressed as a
Kronecker product between the Gram matrix ${G} =
\big(g(x_i,x_j)\big)_{i,j=1}^n$ in $\mathbb{R}^{n \times n}$ and the operator
$T\in \mathcal{L(Y)}$, and is defined as follows:
\begin{equation}
\nonumber
 \mathbf{K} = 
\begin{pmatrix}
g(x_1,x_1)T & \ldots & g(x_1,x_n)T \\
\vdots & \ddots & \vdots \\
g(x_n,x_1)T & \ldots & g(x_n,x_n)T \\
\end{pmatrix} 
= {G} \otimes T.
\end{equation}
It is easy to show that basic properties of the Kronecker product between two 
finite matrices can be restated for this case. So, 
$\mathbf{K}^{-1} = {G}^{-1} \otimes T^{-1}$ and the eigendecomposition of the 
matrix $\mathbf{K}$ can be obtained from the 
eigendecompositions of ${G}$ and $T$ (see Algorithm~\ref{alg:frlsc}). 
\begin{theo}
  \label{th:eigendecomp}
If $T \in \mathcal{L(Y)}$ is a compact, normal operator ($TT^* = T^*T$) on the Hilbert space $\mathcal{Y}$, 
then there exists an orthonormal basis of eigenfunctions $\{\phi_i, i\geq 1\}$ corresponding to 
eigenvalues $\{\lambda_i, i\geq 1\}$ such that 
\begin{equation*}
 T y = \sum\limits_{i=1} \lambda_i \langle y,\phi_i \rangle \phi_i, \quad \forall y\in \mathcal{Y}.
\end{equation*}
\end{theo}

\textbf{Proof}: See~\citet{Naylor-1971}[theorem 6.11.2] \hfill $\blacksquare$
\vspace{2mm}
\newline 
Let $\theta_i$ and ${\mathbf{z_i}}$ be, respectively, the eigenvalues and the eigenfunctions of 
$\mathbf{K}$. From Theorem~\ref{th:eigendecomp} it follows that the inverse operator 
$\mathbf{K}^{-1}$ is given by 
\begin{equation}
 \nonumber
\mathbf{K}^{-1} {\mathbf{c}} = \sum_i \theta_i^{-1} \langle {\mathbf{c}},\mathbf{z_i} \rangle {\mathbf{z_i}}, 
\quad \forall \mathbf{c} \in \mathcal{Y}^n.
\end{equation}
Now we are able to solve the system of linear operator
equations~(\ref{sloe}) and the functions $u_i$ can be computed
from eigenvalues and eigenfunctions of the matrix $\mathbf{K}$, as
described in Algorithm~\ref{alg:frlsc}.
\end{enumerate}
%


%
\begin{algorithm}[tb]
   \caption{$L^2$-Regularized Function-valued Function Learning Algorithm}
   \label{alg:frlsc}
\begin{algorithmic}
   \STATE {\bfseries Input} 
    \\ $\quad$ Examples:
    \\ $\quad \quad$ (function) data $x_i\in (L^2([0,1]))^p$, size $n$
    \\ $\quad \quad$ (function) labels $y_i\in L^2([0,1])$, size $n$
    \\ $\quad$ Parameters: $g$, $T$, $\kappa$, $\lambda$
    \\ \textbf{Eigendecomposition of ${G}$}, the Gram matrix
      of the scalar-valued kernel $g$
    \\ $\quad$ Comment: ${G} =  g(x_i,x_j)_{i,j=1}^n \in \mathbb{R}^{n \times n}$
    \\ $\quad$ Let $\alpha_i \in \mathbb{R}$ the $n$ eigenvalues
    \\ $\quad$ Let $v_i \in \mathbb{R}^n$ the $n$ eigenvectors
    \\ \textbf{Eigendecomposition of} the operator $T \in \mathcal{L}(Y)$
    \\ $\quad$ Choose $\kappa$, the number of computed eigenfunctions
    \\ $\quad$ Compute $\kappa$ $(\delta_i \in \mathbb{R}, 
      w_i \in L^2([0,1]))$ pairs of (eigenvalue, eigenfunction)
    \\ \textbf{Eigendecomposition of $\mathbf{K} = {G} \otimes T$}
    \\ $\quad$ Comment: $\mathbf{K} =  K(x_i,x_j)_{i,j=1}^n \in (\mathcal{L}(Y))^{n \times n}$
    \\ $\quad$ The eigenvalues $\theta_i \in \mathbb{R}$, size $n\times \kappa$,
      are obtained as: $\theta = \alpha \otimes \delta$
    \\ $\quad$ The eigenfunctions $\mathbf{z_i} \in (L^2([0,1]))^n$, size
      $n\times \kappa$, are obtained as: $\mathbf{z} = v \otimes w$
    \\ \textbf{Solution of problem~(\ref{mp1}) $\mathbf{u} = (\mathbf{K}+\lambda I)^{-1} \mathbf{y}$}
    \\ $\quad$ Initialize $\lambda$: regularization parameter
    \\ $\quad$ $\mathbf{u} = \sum_{i=1}^{n\times \kappa} [
    (\theta_i+\lambda)^{-1} \sum_{j=1}^n\langle z_{ij}, y_j \rangle
    \mathbf{z}_{i} ]$
\end{algorithmic}
 \end{algorithm}


To put our algorithm into context, we remind that %
a crucial question about the applicability of functional data is how one can find an appropriate space and a basis in which the functions can be decomposed in a computationally feasible way while taking into account the functional nature of the data. This is exactly what Algorithm~\ref{alg:frlsc} does. In contrast to parametric FDA methods, the basis function here is not fixed in advance but implicitly defined by choosing a \textit{reproducing operator-valued kernel} acting on both input and output data. The spectral decomposition of the block operator kernel matrix naturally allows the assignment of an appropriate basis function to the learning process for representing input and output functions.  Moreover,  the formulation is flexible enough to be used with different operators and then to be adapted for various applications involving functional data.
Also, in the context of nonparametric FDA where the notion of semi-metric plays an important role in modeling functional data, we note that Algorithm~\ref{alg:frlsc} is based on computing and choosing a finite number of eigenfunctions. This is strongly related to the semi-metric building scheme in~\citet{Ferraty-2006} which is based on, for example, functional principal components or  successive derivatives. Operator-valued kernels constructed from the covariance operator~\citep{Kadri-2013} or the derivative operator will allow to design semi-metrics similar to those just mentioned. In this sense, the eigendecomposition of the block operator kernel matrix offers a new way of producing semi-metrics.

\subsection{Generalization Analysis}
\label{sec:generalization}
Here, we provide an analysis of the generalization error of the function-valued function learning model~(\ref{mp}) using the notion of algorithmic stability. For more details and results with the least squares  loss and other loss function~(including $\epsilon$-sensitive loss and logistic loss), see~\citet{Audiffren-ACML2013}. 
In the case of vector-valued functions, the effort in this area has already produced several successful results, including~\citet{Baxter-2000}, \citet{Ando-2005}, \citet{Maurer-2006}, and~\citet{Maurer-2013}.
Yet, these studies have considered only the case of finite-dimensional output spaces, and have focused rather on linear machines than on nonlinear ones. 
To our knowledge, the first work investigating the generalization performance of nonlinear vector-valued function learning methods when output spaces can be infinite-dimensional is that of~\citet{Caponnetto-2006}. In their study, from a theoretical analysis based on the concept of effective dimension, 
the authors have derived generalization bounds for the learning model~(\ref{mp}) when the hypothesis space is an RKHS with operator-valued kernels.

The convergence rates in~\citet{Caponnetto-2006}, although optimal in the case of finite-dimensional output spaces, require assumptions on the kernel that can be restrictive in the infinite-dimensional case. 
Indeed, their proof depends upon the fact that the trace of the operator $K(x,x)$ is finite~($K$ is the operator-valued kernel function), and this restricts the applicability of their results when the output space is infinite-dimensional. To illustrate this, let us consider the identity operator-valued  kernel $K(\cdot,\cdot) = k(\cdot,\cdot) I$, where $k$ is a scalar-valued kernel and $I$ is the identity operator. 
This simple kernel does not satisfy the finite trace condition and therefore the results of~\citet{Caponnetto-2006} cannot be applied in this case. Regarding the examples of operator-valued kernels given in Subsection~\ref{ex}, the kernel built from the integral operator satisfies the finite trace condition, while that based on the multiplication operator does not.
To address this issue, we first show that our learning algorithm is uniformly stable, and then we derive under mild assumption on the kernel,  using a result from~\citet{Bousquet-2002}, a generalization bound which holds even when the finite trace condition is not satisfied. 

We now state and discuss the main assumptions we need to prove a stability-based bound on the generalization error of our method. In the following, we consider a training set $Z=\left\{(x_1,y_1),....,(x_n,y_n) \right\}$ of size $n$ in $\x \times \y$ drawn i.i.d. from an unknown distribution~$P$, and we denote by $\zi =Z \setminus (x_i,y_i)$ the set $Z$ from which the couple $(x_i,y_i)$ is removed.
We will use a cost function $c:\y \times \y \to \mathbb{R}_+$. The loss of an hypothesis $F$ with respect to an example $(x,y)$ is then defined as $\ell(y,F,x) = c(F(x),y)$. 
The generalization error is defined as:
$$R(F) = \int \ell(y,F(x),x)  dP(x,y),$$ 
and the empirical error as:
$$R_{emp}(F,Z)=\frac{1}{n}\sum_{i=1}^n \ell(y_i,F,x_i).$$
A learning algorithm can be viewed as a function which maps a training set $Z$ onto a function $F_Z$ from $\x$ to $\y$~\citep{Bousquet-2002}. In our case, $F_Z$ is the solution of the optimization problem~(\ref{mp}) which is an instance of the following scheme
\begin{equation}
\label{definition fZ}
F_Z = \arg\min_{F \in \mathcal{F}} R_{reg}(F,Z),
\end{equation}
where $R_{reg}(F,Z) = R_{emp}(F,Z) +  \lambda \| F \|_\mathcal{F}^2$.

\begin{hypothesis}\label{Hyp K bounded}
$\exists \kappa >0$ such that $\forall x \in \x$, $$\| K(x,x) \|_{op} \le \kappa ^ 2,$$
where $\displaystyle\| K(x,x) \|_{op}= \sup_{y \in \y} \frac{\|K(x,x)y\|_\y} {\|y\|_\y}$ is the operator norm of $K(x,x)$ on $L(\y)$.
\end{hypothesis}
\begin{hypothesis}\label{Measure}
The real function from $\x \times \x \to \R$
$$(x_1,x_2) \mapsto \left\langle K(x_1,x_2)y_1,y_2 \right\rangle_\y \text{ is measurable } \forall y_1,y_2\in \y.$$
\end{hypothesis}
\begin{hypothesis} \label{Hyp on c}
The application $(y,f,x)\mapsto \ell(y,F,x) $ is $\sigma$-admissible, i.e.  convex with respect to $F$ and Lipschitz continuous with respect to $F(x)$, with $\sigma$ its Lipschitz constant.
\end{hypothesis}
\begin{hypothesis}\label{c control}
$\exists \xi>0$ such that $\forall (x,y) \in \x \times \y$ and $\forall Z$ a training set,
$$\ell(y,F_Z,x)\le \xi.$$
\end{hypothesis}
Note that Assumption~\ref{Hyp K bounded} is a direct extension from the scalar-valued to the operator-valued case of the boundedness condition of the kernel function. It replaces and weakens the finite trace assumption of the operator $K(x,x)$ used in~\citet{Caponnetto-2006}; see Remark~\ref{remarkHS} for more details. Assumption~\ref{Measure} was also used by~\citet{Caponnetto-2006} to avoid problems with measurability. This assumption with the fact that $\mathcal{F}$ is separable 
ensures that all functions in $\mathcal{F}$ are measurable from $\x$ to $\y$. Assumptions~\ref{Hyp on c} and~\ref{c control} are the same as those used by~\citet{Bousquet-2002} for learning scalar-valued functions. 
As a consequence of Assumption \ref{Hyp K bounded}, we immediately obtain the following elementary lemma which allows to control $\|F(x)\|_\y$ with $\|F\|_\mathcal{F}$.
\begin{lemma}\label{Lem reprod}
Let $K$ be a nonnegative operator-valued kernel satisfying Assumption \ref{Hyp K bounded}. Then $\forall F \in \mathcal{F}$, $\|F(x)\|_\y \le \kappa \|F\|_{\mathcal{F}} $.
\end{lemma}
\medskip
\textbf{Proof}:
\begin{align*}
  \begin{array}{ll}
    \|F(x)\|_{\mathcal{Y}} &= \sup\limits_{\|y\|=1} | \langle F(x), y \rangle _{\mathcal{Y}}|
    =   \sup\limits_{\|y\|=1} | \langle F(\cdot),K(x,\cdot) y \rangle _{\mathcal{F}} | \\[0.5cm]
    & \leq \|F(\cdot)\|_{\mathcal{F}} \sup\limits_{\|y\|=1} \|K(x,\cdot) y \|_{\mathcal{F}}
    \leq \|F(\cdot)\|_{\mathcal{F}} \sup\limits_{\|y\|=1} \sqrt{\langle K(x,x) y, y \rangle_{\mathcal{Y}}} \\[0.5cm]
     & \leq \|F(\cdot)\|_{\mathcal{F}}  \sup\limits_{\|y\|=1} \|K(x,x) y\|^{\frac{1}{2}}_{\y} \leq \|F(\cdot)\|_{\mathcal{F}} \|K(x,x) \|^{\frac{1}{2}}_{op} 
      \le \kappa \|F\|_{\mathcal{F}}  
  \end{array}
\end{align*}
     \hfill $\blacksquare$

Now we are ready to state the stability theorem for our function-valued function learning algorithm. This result is a straightforward extension of \hbox{Theorem 22} in~\citet{Bousquet-2002} to the case of infinite-dimensional output spaces. It is worth pointing out that the proof does not differ much from the scalar-valued case and requires only minor modifications to fit the operator-valued kernel approach. For the convenience of the reader, we present in Appendix~\ref{app:stability}  the proof taking into account these modifications. 
Before stating the theorem we would like to recall the definition of uniform algorithmic stability from~\citet{Bousquet-2002}.
\begin{de} 
A learning algorithm $Z \mapsto F_Z$ has uniform stability $\beta$ with respect to the loss function $\ell$ if the following holds
$$\forall n \ge 1, \text{ } \forall 1 \le i \le n, \text{ } \forall Z \text{ a training set, \ } \|\ell(\cdot,F_Z,\cdot) - \ell(\cdot,F_\zi,\cdot)\|_{\infty} \le \beta$$
\end{de}
\begin{theo} \label{beta stable}
Under Assumptions \ref{Hyp K bounded}, \ref{Measure} and \ref{Hyp on c}, a learning algorithm that maps a training set $Z$ to the function $F_Z$ defined in~(\ref{definition fZ}) is $\beta$ stable with $$\beta= \frac{\sigma^2 \kappa^2 }{2\lambda n}.$$
\end{theo}
\textbf{Proof}: See Appendix \ref{app:stability}.      \hfill $\blacksquare$\\[0.4cm]
$\beta$ scales as $1/n$. This allows to get a bound on the generalization error using a result from~\citet{Bousquet-2002}.
\begin{theo}\label{consistence}
Let $Z \mapsto F_Z$ be a learning algorithm with uniform stability $\beta$ with respect to a loss  $\ell$ that satisfies Assumption~\ref{c control}. Then, $\forall n \ge 1$, $\forall \text{ }0 \le \delta\le 1$, the following bound holds with probability at least $1-\delta$ over the random draw of training samples
\begin{equation}\nonumber
\begin{aligned}
R \le R_{emp} + 2\beta+(4 n \beta + \xi)\sqrt{\frac{\ln(1/\delta)}{2n}}.
\end{aligned}
\end{equation}
\end{theo}
\textbf{Proof}: See Theorem~12 in~\citet{Bousquet-2002}. \hfill $\blacksquare$
\bigskip

For our learning model~(\ref{mp}), we should note that Assumption~\ref{Hyp on c} is in general not satisfied with the least squares loss function $\ell(y,F,x)= \|y - F(x) \|_\y^ 2$. To address this issue, one can add a boundedness assumption on $\y$, which is a sufficient condition to prove the uniform stability when Assumption~\ref{Hyp K bounded} is satisfied.
\begin{hypothesis}\label{Hyp Y bounded}
$\exists \sigma_y>0$ such that $\|y\|_\y<\sigma_y$, $\forall y \in \y$.
\end{hypothesis}
\begin{lemma}\label{Hyp LSR} 
Let $\ell(y,F,x)= \|y - F(x) \|_\y^ 2$. If Assumptions~\ref{Hyp K bounded} and~\ref{Hyp Y bounded} hold, then
$$\vert \ell(y,F_Z,x) -  \ell(y,F_\zi,x)\vert \le \sigma \|F_Z(x)-F_\zi(x)\|_\y ,$$
with $\displaystyle \sigma=2 \sigma_y ( 1 + \frac{\kappa}{\sqrt{\lambda}})$.
\end{lemma}
\textbf{Proof}: See Appendix~\ref{app:admissible}.      \hfill $\blacksquare$\\[0.3cm]
This Lemma can replace the Lipschitz property of $\ell$ in the proof of Theorem~\ref{beta stable}.
Moreover, Assumptions~\ref{Hyp K bounded} and~\ref{Hyp Y bounded} are sufficient to satisfy Assumption~\ref{c control} with $\xi= (\sigma/2)^2$~(see Appendix~\ref{app:admissible}).  
We can then use Theorem~\ref{beta stable} to prove the uniform stability of our function-valued function learning algorithm with $$\beta=\displaystyle\frac{2 \kappa^2 \sigma^2_y  ( 1 + \displaystyle\frac{\kappa}{\sqrt{\lambda}})^2 }{\lambda n}.$$
Theorem~\ref{consistence} thus gives us a bound on the generalization error of our method equal, with probability at least $1- \delta$, to
\begin{equation}\nonumber
\begin{aligned}
R \le R_{emp}&+\frac{4 \kappa^2  \sigma^2_y (1 + \displaystyle\frac{\kappa}{\sqrt{\lambda}})^2 }{\lambda n}
+\sigma^2_y ( 1 + \frac{\kappa}{\sqrt{\lambda}})^2( \frac{8  \kappa^2}{\lambda} + 1)\sqrt{\frac{\ln(1/\delta)}{2 n}} . 
\end{aligned}
\end{equation}

\begin{rem}
\label{remarkHS}
It is important to stress that even though the stability analysis of function-valued function learning algorithms follows in a quite straightforward fashion from the earlier results presented in~\citet{Bousquet-2002} and provides convergence rates which are not optimal, it allows to derive generalization error bounds with operator-valued kernels for which the trace of the operator $K(x,x)$ is not necessarily finite.
Assumption~\ref{Hyp K bounded} is weaker than the one used in~\citet{Caponnetto-2006} which requires that the operator $K_x$ is Hilbert-Schmidt\footnote{The operator $K_x$ from $\y$ to $\mathcal{F}$, defined by $y \mapsto K(x,\cdot)y$, $\forall y \in \y$, is a Hilbert-Schmidt operator if, for some any basis $(y_j)_{j\in \N}$ of $\y$, it holds that $Tr(K_x^*K_x) = \sum_j  \left\langle K(x,\cdot)y_j,K(x,\cdot)y_j \right\rangle_\mathcal{F}  < +\infty$. This is equivalent to saying that the operator $K(x,x) \in \mathcal{L(Y)}$ is of trace class, since by the reproducing property we have  $\left\langle K(x,\cdot)y_j,K(x,\cdot)y_j \right\rangle =  \left\langle K(x,x)y_j,y_j \right\rangle_\y$.} and $\sup_{x \in \x} Tr(K(x,x)) <  \kappa $.
 While the two assumptions are equivalent when the output space $\y$ is finite dimensional, this is no longer the case when, as in this paper,  $\dim\y=+\infty$.
 Moreover, we observe that if the assumption of~\citet{Caponnetto-2006} is satisfied,  then our Assumption~\ref{Hyp K bounded} holds (see proof in Appendix~\ref{app:HS}). The converse is not true (see Remark~\ref{rem:counterexample} for a counterexample).
\end{rem}

\begin{rem}
\label{rem:counterexample}
Note that the operator-valued kernel based on the multiplication operator and described in Subsection~\ref{ex} satisfies Assumption~\ref{Hyp K bounded} but not the finite trace condition as assumed in~\citet{Caponnetto-2006}.
 Let $k$ be a positive-definite scalar-valued kernel such that $\sup_{x \in \x} k(x,x) < +\infty$, $\I$ an interval of $\R$, $\mu>0$, and $\y~=~L^2(\I,\R)$. Let $ f \in L^ \infty (\I,\R)$ be such that $\|f\|_\infty< \mu$.
Consider the following  multiplication operator-valued kernel $K$:
$$K(x,z)y(\cdot)= k(x,z) f^2(\cdot) y(\cdot) \in \y.$$ 
$K$ is a nonnegative operator-valued kernel. While $K$ always satisfies Assumption~\ref{Hyp K bounded}, the Hilbert-Schmidt property of $K_x$ depends on the choice of $f$ and does not hold in general. For instance, let $f(t)= \displaystyle\frac{\mu}{2} (\exp(-t^2) +1)$, then 
$$ \| K(x,x) \|_{op} \le \mu^2 k(x,x), $$
and 
$$ Tr(K(x,x))= \sum_{j \in \N} \left\langle K(x,x) y_j,y_j \right\rangle  \ge k(x,x) \displaystyle\frac{\mu}{2} \sum_{i \in \N} \|y_j\|_2^2 = \infty, $$
where $(y_j)_{j \in \N}$ is an orthonormal basis of $\y$ (which exists since $\y$ is separable).
\end{rem}


\section{Experiments}
\label{sec:exp}

In this experimental section, we essentially aim at illustrating the
potential of adopting a functional data analysis perspective for
learning multi-output functions when the data are curves. First, we
are interested in the problem of acoustic-to-articulatory speech
inversion where the goal is to learn vocal tract (VT) time functions
from the acoustic speech signal~\citep{Mitra-2010}. Then we show,
through experiments on sound recognition~\citep{Asma-2008}, that the
proposed framework can be applied beyond functional response
regression, for problems like multiple functional classification where
each sound to be classified is represented by more than one functional
parameters.

The operator-valued kernel used in these experiments is the kernel $K$
defined by Equation\@~\eqref{kernn}. We use the inner product in
$\mathcal{X}^p$ for the scalar-valued kernel $g$, where $p$ is~the 
number of functional parameters of a speech or a sound
signal. Also, extending real-valued functional kernel, as
in~\citet{Rossi-2006}, to multiple functional inputs could be
possible.  Eigenvalues $\delta_i$ and eigenfunctions $w_i$ of the
Hilbert-Schmidt integral operator $T$ associated with the
operator-valued kernel $K$ are equal to $\frac{2}{1+\mu_i^2}$ and
$\mu_i\cos(\mu_ix)+\sin(\mu_ix)$ respectively, where $\mu_i$ are
solutions of the equation $\cot \mu = \frac{1}{2}(\mu-\frac{1}{\mu})$.
Eigendecomposition of an infinite dimensional operator $T$ is computed
in general by solving a differential equation obtained from the
equality $Tw_i=\delta_i w_i$.

In order to choose the regularization parameter $\lambda$ and the
number of eigenfunctions $\kappa$ that guarantee optimal solutions,
one may use the cross-validation score based on the
one-curve-leave-out prediction error~\citep{Rice-1991}.  Then we
choose $\lambda$ and $\kappa$ so as to minimize the cross-validation
score based on the squared prediction error
\begin{equation}
 CV(\lambda) = \sum\limits_{i=1}^n \sum\limits_{j=1}^{N_i}\{ y_{ij} - \hat{y}_i^{(-i)}(t_{ij})\}^2,
\end{equation}
where $n$ is the number of functions $y_i$, $y_{ij}$ is the observed value at time $t_{ij}$, 
$N_i$ the number of measurements made on $y_i$ and $\hat{y}_i^{(-i)}$ the predicted curve 
for the $i^{\mbox{\scriptsize{}th}}$ function, computed after removing the data for this function.

\subsection{Speech Inversion}

\begin{figure}[t]
  \centering
  \includegraphics[scale=0.18]{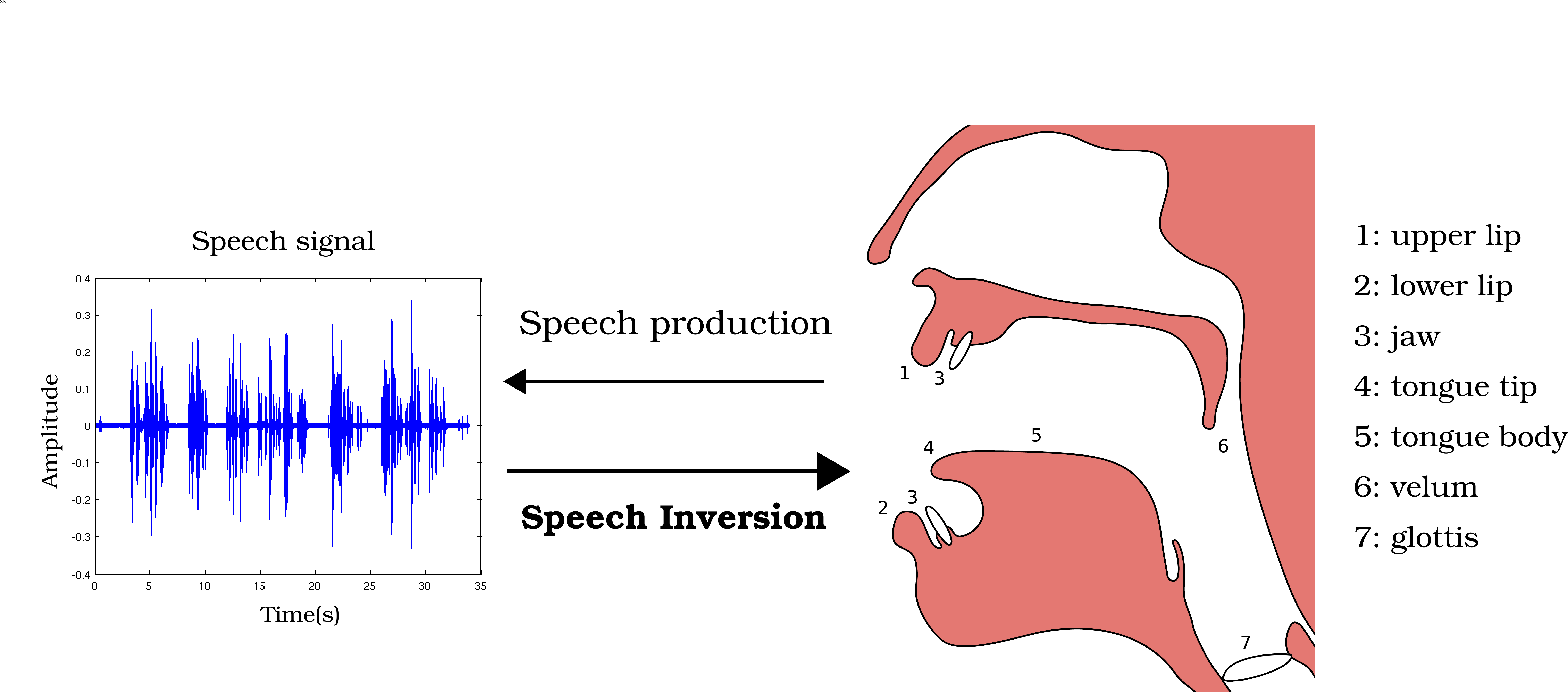}
  \caption{Principle of speech inversion (a.k.a.
    acoustic-articulatory inversion). Human beings produce an audible
    speech signal by moving their articulators (\eg{} tongue, lips,
    velum, etc.) to modify a source of sound energy in the
    vocal tract.
    In performing the inversion mapping, we aim to invert this forward
    direction of speech production.  In other words, we aim to take a
    speech signal and estimate the underlying articulatory movements
    which are likely to have created it~\citep{Richmond-2002}.  }
\label{SpeechInversion}
\end{figure}

The problem of speech inversion has received increasing attention in
the speech processing community in the recent years
(see~\citet{Schroeter-1994, Mitra-2010, Kadri-2011a} and references
therein).  This problem, aka acoustic-articulatory inversion, involves
inverting the forward process of speech production (see
Figure~\ref{SpeechInversion}). In other words, for a given acoustic
speech signal we aim at estimating the underlying sequence of
articulatory configurations which produced it~\citep{Richmond-2002}.
Speech inversion is motivated by several applications in which it is
required to estimate articulatory parameters from the acoustic speech
signal. For example, in speech recognition, the use of articulatory
information has been of interest since speech recognition efficiency
can be significantly improved~\citep{Kirchoff-1999}.  This is due to
the fact that automatic speech recognition (ASR) systems suffer from
performance degradation in the presence of noise and spontaneous
speech. Moreover, acoustic-to-articulatory speech inversion is also
useful in many other interesting applications such as speech analysis
and synthesis~\citep{Toda-2004} or helping individuals with speech and
hearing disorders by providing visual feedback~\citep{Toutios-2005b}.

 \setlength\textfloatsep{8pt plus 2pt minus 2pt}

Most of current research on acoustic-to-articulatory inversion focuses
on learning Electromagnetic Articulography (EMA) trajectories from
acoustic parameters and frequently uses the MOCHA \texttt{fsew0}
data set as training and test data~\citep{Richmond-2002}.  In a recent
work,~\citet{Mitra-2010} suggest the use of the TAsk Dynamics
Application (TADA) model~\citep{TADA-2004} to generate
acoustic-articulatory database which contains synthetic speech and the
corresponding vocal tract time functions.  Their results show
that tract variables can be better candidates than EMA trajectories
for articulatory feature based ASR systems. In our experiments, we
follow this work by addressing the issue of finding the mapping
between acoustic parameters and vocal tract variables. In this
context, we use Mel-Frequency Cepstral Coefficients (MFCCs) as input
and consider as output eight different vocal tract constriction
variables, lip aperture (LA), lip protrusion (LP), tongue tip
constriction degree (TTCD), tongue tip constriction location (TTCL),
tongue body constriction degree (TBCD), tongue body constriction
location (TBCL), Velum (VEL) and Glottis (GLO). Table~\ref{VTV} shows
the eight vocal tract variables we used in this study and the
corresponding constriction organs and articulators~\citep{Mitra-2009}.

\begin{figure}[p]
\centering
\includegraphics[scale=0.95]{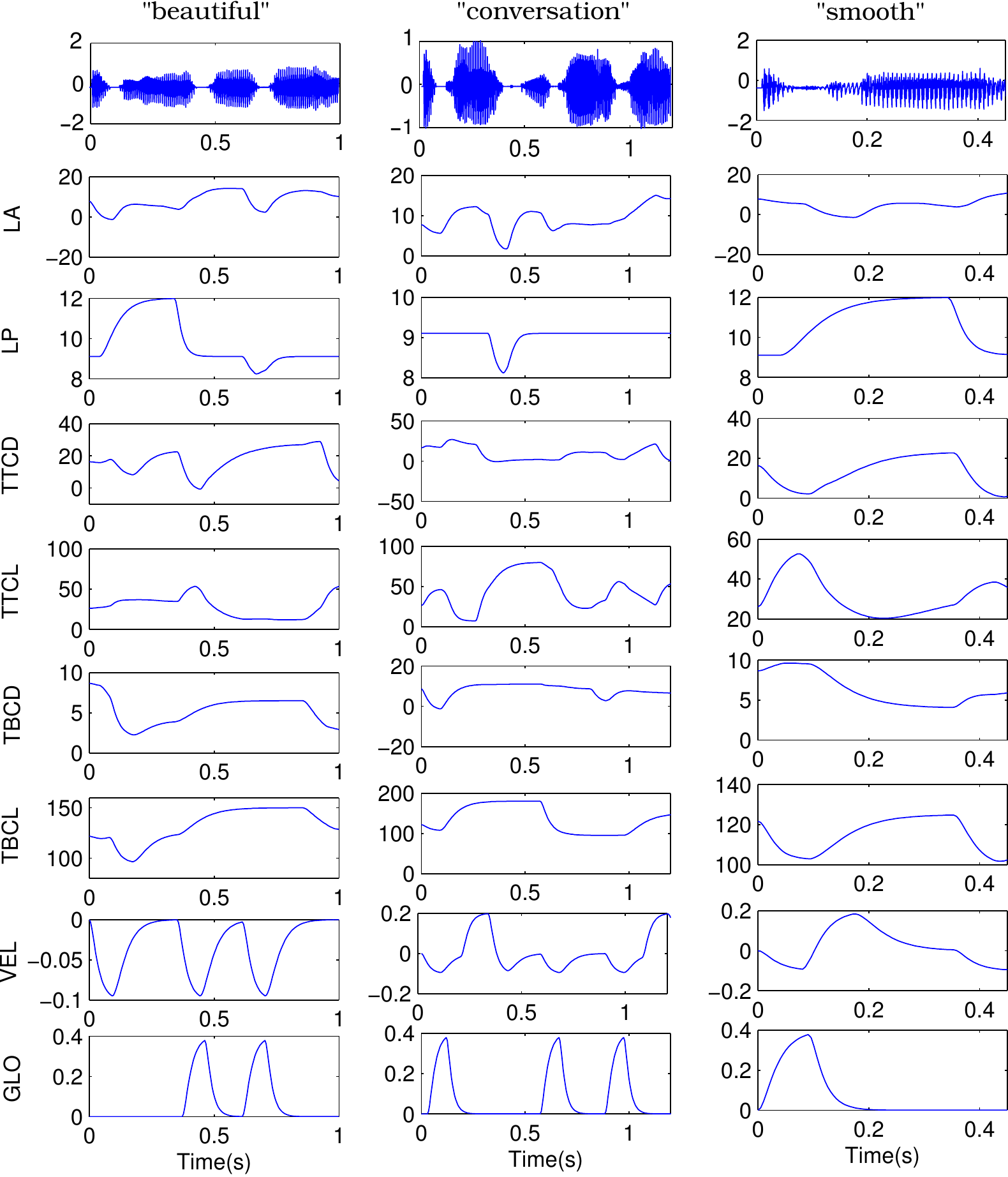} 
\caption{Acoustic waveforms and derived vocal tract time functions for
  the utterances ``beautiful'', ``conversation'' and ``smooth''.  The
  vocal tract variables are: lip aperture (LA), lip protrusion (LP),
  tongue tip constriction degree (TTCD), tongue tip constriction
  location (TTCL), tongue body constriction degree (TBCD), tongue body
  constriction location (TBCL), Velum (VEL) and Glottis (GLO).}
\label{vttf}
\end{figure}

Moreover, articulators move relatively slowly and smoothly, and their
movements are continuous. Indeed, the mouth cannot ``jump'' from one
configuration to a completely different one~\citep{Richmond-2002}. For
this reason, functional data analysis approaches are well suited for
the speech inversion task. In other words, even if the measurement
process itself is discrete, vocal tract variables are really smooth
functions (see Figure~\ref{vttf}) rather than vectors and taking into
account such prior knowledge on the nature of the data can
significantly improve performance.
In our proposed method, smoothness is guaranteed by the use of smooth
eigenfunctions obtained from the spectral decomposition of the
integral operator associated with a Mercer kernel used to construct
the operator-valued kernel defined in Equation~\ref{kernn}.
By this way, our approach does not need the filtering post-processing
step, which is necessary in vectorial vocal-tract learning methods to
transform the predicted functions on smooth curves and which has the
drawback of %
changing the behavior of the predicted vocal tract time functions.

\begin{table}[t]
\renewcommand{\arraystretch}{1.3}
\begin{center}
\small
\begin{tabular}{|>{\centering}p{3.8cm}|>{\centering}p{4.4cm}|>{\centering}p{3.9cm}|}
            \hline
              \textbf{Constriction organ} & \textbf{VT variables}   &  \textbf{Articulators} \tabularnewline
              \hline 
              \multirow{2}{*}{lip}        & lip aperture (LA)    &  \multirow{2}{*}{upper lip, lower lip, jaw}   \tabularnewline \cline{2-2} 
                  & lip protrusion (LP)       &     \\ \hline
              \multirow{4}{*}{tongue tip} & tongue tip constriction degree (TTCD)   & \multirow{4}{*}{tongue body, tip, jaw}  \tabularnewline \cline{2-2}
                     &  tongue tip constriction location (TTCL)       &       \tabularnewline \hline
              \multirow{4}{*}{tongue body}  & tongue body constriction degree (TBCD)  & \multirow{4}{*}{tongue body, jaw}   \tabularnewline \cline{2-2}
                         & tongue body constriction location (TBCL)      &    \tabularnewline \hline
	      velum              & velum (VEL)          &    velum   \tabularnewline \hline
                glottis         & glottis (GLO)      & glottis   \tabularnewline
	      \hline
\end{tabular}
\end{center}
\caption{Constriction organ, vocal-tract~(VT) variables and involved articulators~\citep{Mitra-2009}. \vspace{-0.2cm}}
\label{VTV}
\end{table}

Various nonlinear acoustic-to-articulatory inversion
techniques~\citep{Richmond-2002, Mitra-2010}, and particularly
kernel-based methods~\citep{Toutios-2005b, Mitra-2009}, have been
proposed in the literature. In most cases, these works address the
articulatory estimation problem within a single-task learning
perspective. However, in~\citet{Richmond-2007a} and more recently
in~\citet{Kadri-2011a}, the authors put forward the idea that we can
benefit from viewing the acoustic-articulatory inversion problem from a
multi-task learning perspective.
Motivated by comparing our functional operator-valued kernel based
approach with multivariate kernel methods, we report on experiments
similar to those performed by~\citet{Mitra-2009}
and~\citet{Kadri-2011a}.
The tract variables learning technique proposed by~\citet{Mitra-2009}
is based on a hierarchical $\varepsilon$-SVR architecture constructed
by associating different SVRs, a SVR for each tract variable. To
consider the dependencies between VT time functions, the SVRs
corresponding to independent VT variables are first created and then
used for constructing the others. Otherwise, the
acoustic-to-articulatory method in~\citet{Kadri-2011a} is based on
learning a vector-valued function using a matrix-valued kernel
proposed in~\citet{Caponnetto-2008}.

Following~\citet{Mitra-2010}, acoustic-articulatory database is
generated by the TADA model~\citep{TADA-2004} which is a computational
implementation of articulatory phonology. The generated data set
consists of acoustic signals for 416 words chosen from the Wisconsin
X-ray microbeam data~\citep{Microbeam-1994} and corresponding Vocal
Tract (VT) trajectories sampled at 5 ms.  The speech signal was
parameterized into 13 Mel-Frequency Cepstral Coefficients (MFCC).
These Cepstral coefficients were acquired each 5 ms
(synchronized with the TVs) with window duration of 10 ms.

\begin{table}[t]
  \begin{center}
    \begin{tabular}{cccc}\hline
    {VT variables}   &  {$\varepsilon$-SVR} & {Multi-task}  & Functional \\ 
\hline
\hline
    LA & 2.763 &  2.341 & \textbf{1.562} \\
    LP &  0.532 & \textbf{0.512}& 0.528\\
    TTCD &  3.345 & 1.975 & \textbf{1.647} \\
    TTCL &  7.752 & 5.276 & \textbf{3.463} \\
    TBCD &  2.155 & 2.094 & \textbf{1.582} \\
    TBCL & 15.083 & 9.763 & \textbf{7.215} \\
    VEL &  0.032  & 0.034 & \textbf{0.029} \\
    GLO &  \textbf{0.041} & 0.052 & 0.064 \\\hline\hline
  Total  &  3.962      & 2.755 & \textbf{2.011}\\
  \hline
    \end{tabular}
  \end{center}
 \caption{Average RSSE for the tract variables using hierarchical 
$\varepsilon$-SVR~\citep{Mitra-2009}, the multi-task kernel method~\citep{Kadri-2011a} 
and the proposed functional operator-valued kernel based approach.}
  \label{tab}
\end{table}

For evaluating the performance of the VT time functions estimation,
we use the residual sum of squares error (RSSE) defined as follows
\begin{equation}
 \label{rss}
    RSSE = \int\sum\limits_{i}\{y_i(t)-\widehat{y}_i(t)\}^2 dt,
\end{equation}
where $\widehat{y}_i(t)$ is the prediction of the VT curve $y_i(t)$.
Table~\ref{tab} reports average RSSE results obtained using the
hierarchical $\varepsilon$-SVR algorithm~\citep{Mitra-2009}, the
multi-task kernel method~\citep{Kadri-2011a} after smoothing the
estimated VT trajectories using a Kalman filter as described
in~\citet{Mitra-2009}, and the functional operator-valued kernel based
approach.
The proposed functional approach consistently produced significant
performance improvements over the supervised baseline
$\varepsilon$-SVR. It also outperforms the discrete multi-task
method~\citep{Evgeniou-2005,Kadri-2011a} except for the LP and GLO
variables. The multi-task and also the $\varepsilon$-SVR methods
perform well for these two vocal tract variables and slightly improve our
functional approach.  This can be explained by the fact that, contrary
to other vocal tract variables, LP and GLO time functions are not
completely smooth for all times and positions, while our method with
the integral operator-valued kernel, as defined in
Equation~\ref{kernn}, tends to favor the prediction of smooth
functions. Building operator-valued kernels suitable for
heterogeneous functions, i.e., smooth in some parts and non-smooth in
others, could be a good alternative to improve the prediction of these
two vocal tract time functions.
Note that the number of eigenfunctions $\kappa$ affects performance. $\kappa$ has to be well 
chosen to provide a reasonable approximation of the infinite-dimensional process. 
In the case of complex output functions, like heterogeneous functions, 
we need to use many eigenfunctions to have a good approximation, but even for this case, 
$\kappa$ remains (very) small compared to the number of examples~$n$.

\subsection{Sound Recognition}

A second application of the ideas that we present in this paper is sound recognition.
Many previous works in the context of sound recognition problem have concentrated on classifying environmental
sounds other than speech and music~\citep{Dufaux-2000,Peltonen-ICASSP-2002}. Such
sounds are extremely versatile, including signals generated in domestic, business, and outdoor environments. 
A system that is able to recognize such sounds may be of great importance for surveillance
and security applications~\citep{Istrate-2006,Asma-2008}. 
The classification of a sound is usually performed in two steps.
First, a pre-processor applies signal processing techniques to
generate a set of features characterizing the signal to be
classified. Then, in the feature space, a decision rule is implemented to assign a class to a pattern.

Operator-valued kernels can be used in a classification setting by
considering the labels $y_i$ to be functions in some function space
rather than real values. Similarly to the scalar case, a natural
choice for $y_i$ would seem to be the Heaviside step function in
$L^2([0,1])$ scaled by a real number. In this context, our method can
be viewed as an extension of the Regularized Least Squares
Classification (RLSC) algorithm~\citep{Rifkin-2003} to the FDA domain
(we called it Functional RLSC~\citep{Kadri-2011c}). The performance of
the proposed algorithm described in Section~\ref{fvfe} is evaluated on
a data set of sounds collected from commercial databases which include
sounds ranging from screams to explosions, such as gun shots or glass
breaking, and compared with the RLSC method.

\begin{table}[t]
\begin{center}
\begin{tabular}{cccccc}
  \hline
  Classes & Number &Train & Test & Total  & Duration (s)\\
  \hline\hline
Human screams   & C1    & 40    & 25  & 65    &167\\
Gunshots        & C2    & 36    & 19  & 55    & 97\\
Glass breaking  & C3    & 48    & 25  & 73    &123\\
Explosions      & C4    & 41    & 21  & 62    &180\\
Door slams      & C5    & 50    & 25  & 75    & 96\\
Phone rings     & C6    & 34    & 17  & 51    &107\\
Children voices & C7    & 58    & 29  & 87    &140\\
Machines        & C8    & 40    & 20  & 60    &184\\\hline\hline
  Total         &       & 327   & 181 & 508   &18mn 14s \\
  \hline
\end{tabular}
\end{center}
\caption{Classes of sounds and number of samples in the database used for performance evaluation.\label{tab:1}}
\end{table}

\subsubsection{Database Description}
As in~\cite{Asma-2008}, the major part of the sound samples used in the recognition
experiments is taken from two sound libraries~\citep{Leonardo-Software,Real-World}. 
All signals in the database have a 16 bits
resolution and are sampled at 44100 Hz, enabling both good time
resolution and a wide frequency band, which are both necessary to
cover harmonic as well as impulsive sounds. 
The selected sound classes are given in Table~\ref{tab:1}, and they are typical
of surveillance applications. The number of items in each class is
deliberately not equal.

Note that this database includes impulsive and harmonic sounds
such as phone rings (C6) and children voices (C7).  These sounds are
quite likely to be recorded by a surveillance system.  Some sounds are
very similar to a human listener: in particular, explosions (C4) are
pretty similar to gunshots (C2).  Glass breaking sounds include both
bottle breaking and window breaking situations.  Phone rings are
either electronic or mechanic alarms.

\begin{figure}[p]
  \centering
\includegraphics[height=0.26\textheight, width=0.7\columnwidth]{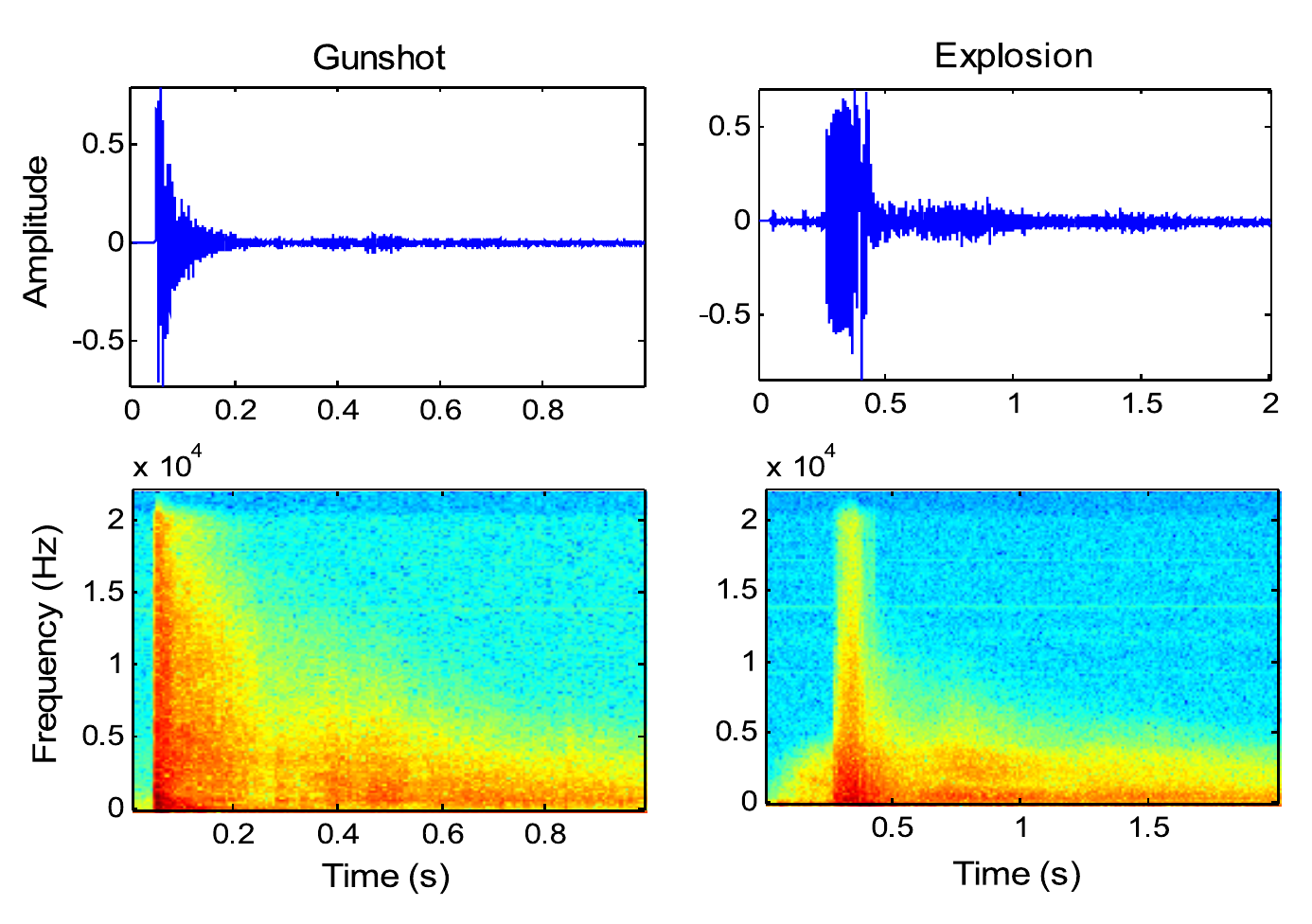}

  \caption{Structural similarities between two different classes. 
Gunshot and Explosion are two sound signals belonging to two different classes, but they have similar 
temporal and spectral representations.}
\label{similarity}
\end{figure}

\begin{figure}[p]
  \centering
\includegraphics[scale=0.8]{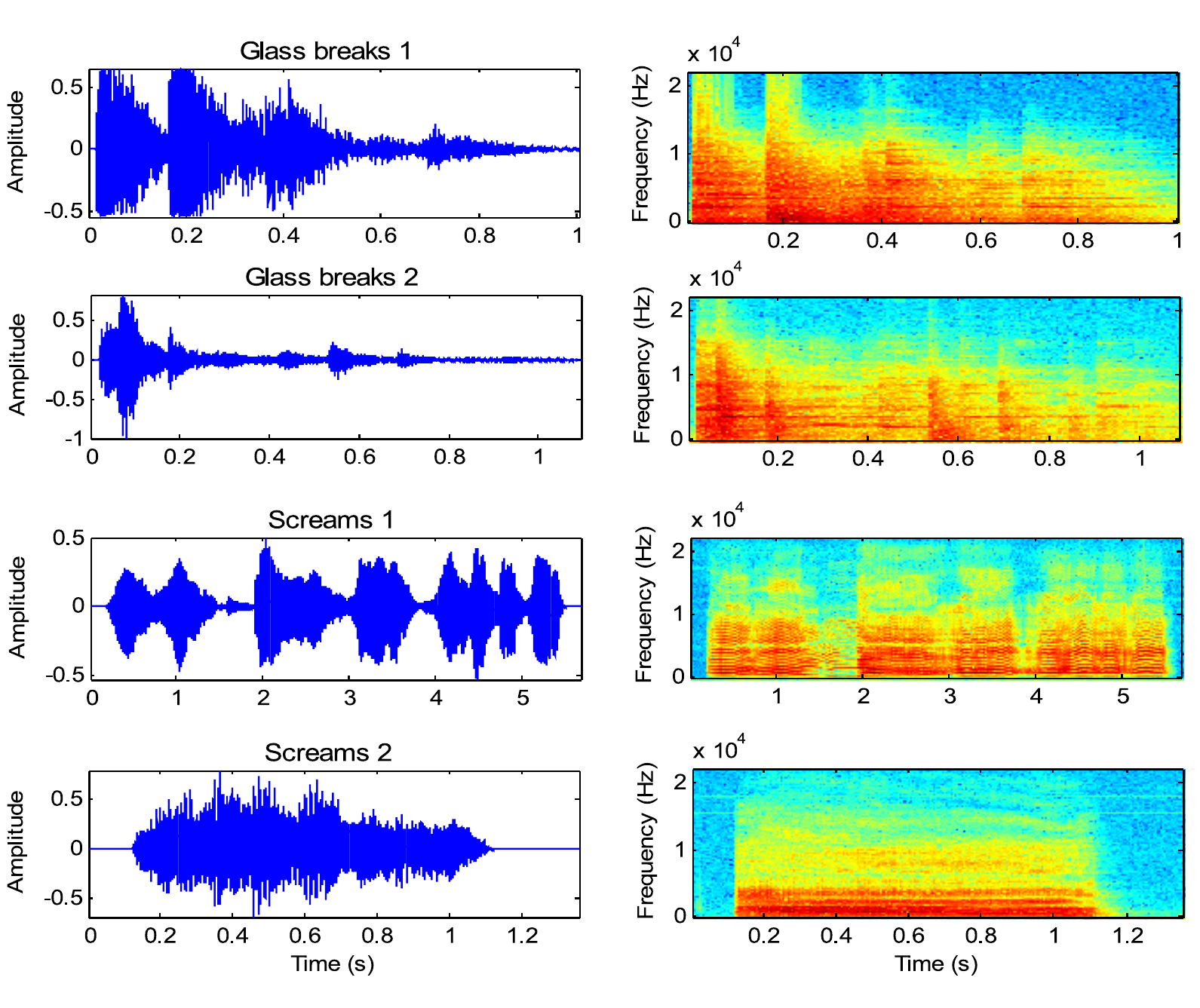}
  \caption{Structural diversity inside the same sound class and between classes.
Glass~breaks~1~and~2 (resp.~Screams~1~and~2) are two sounds from the same class, however they 
present different temporal (resp.~spectral) properties.}
\label{structure-temporel}
\end{figure}

Temporal representations and spectrograms of some sounds are depicted
in Figures~\ref{similarity} and~\ref{structure-temporel}.  Power
spectra are extracted through the Fast Fourier Transform (FFT) every
10 ms from 25 ms frames.  They are represented vertically at the
corresponding frame indexes.  The frequency range of interest is
between 0 and 22 kHz.  A lighter shade indicates a higher power value.
These figures show that in the considered database we can have both:
(1) many similarities between sounds belonging to different classes,
(2) diversities within the same class of sounds.
%

%
\subsubsection{Sound Classification Results}


Following~\cite{Rifkin-2004}, the 1-vs-all multi-class classifier is
selected in these experiments.  So we train $N$ (number of classes)
different binary classifiers, each one trained to distinguish the data
in a single class from the examples in all remaining classes. We run
the $N$ classifiers to classify a new example.

The adopted sound data processing scheme is the following.~Let
$\mathcal{X}$ be the set of training sounds, shared in $N$ classes
denoted ${\mathcal{C}}_1,\ldots,{\mathcal{C}}_N$. Each class contains
$m_i$ sounds, $i=1,\ldots,N$. Sound number $j$ in class
${\mathcal{C}}_i$ is denoted ${\mathbf s}_{i,j}$, ($i=1,\ldots,N,
j=1,\ldots,m_i$).  The pre-processor converts a recorded acoustic
signal ${\mathbf s}_{i,j}$ into a time/frequency localized
representation.
In multivariate methods, this representation is obtained by splitting
the signal ${\mathbf s}_{i,j}$ into $T_{i,j}$ overlapping short frames
and computing a vector of features $z_{t,i,j}$, $t=1,\ldots,T_{i,j}$
which characterize each frame.
Since the pre-processor is a series of continuous time-localized
features, it will be useful to take into account the relationships
between feature samples along the time axis and consider dependencies
between features. That is why we use a FDA-based approach in which
features representing a sound are modeled by functions $z_{i,j}(t)$.
In this work, Mel Frequency Cepstral Coefficients (MFCCs) features are
used to describe the spectral shape of each signal. These coefficients
are obtained using 23 channels Mel filterbank and a Hamming analysis
window of length $25$ ms with $50\%$ overlap. We also use the energy
parameter measured for each window along all the sound signal. So,
each sound is characterized by 14 functional parameters: 13 Cepstral
functions and 1 energy function.
Performance of the proposed functional approach in the context of
classification is compared to the results obtained by the RLSC
algorithm, see Tables~\ref{tab2} and~\ref{tab1}.  The performance is
measured as the percentage number of sounds correctly recognized and
it is given by $(\mathrm{W_r}/\mathrm{T_n})\times100\%$, where
$\mathrm{W_r}$ is the number of well recognized sounds and
$\mathrm{T_n}$ is the total number of sounds to be recognized.
The use of the Functional RLSC is fully justified by the results presented here, as it yields consistently 
a high classification accuracy for the major part of the sound classes.

RLSC setup is similar to that of FLRSC. The major difference is in the modeling of sound features. 
In RLSC, all the functional parameters which characterize a sound are combined in the same vector which is 
considered to be in $\mathbb{R}^d$. Functional RLSC considers each input sound as a vector of functions in 
$(L^2([0,1]))^p$ where $p$ is the number of functional parameters. By using operator-valued kernels rather than 
scalar-valued ones, we project these functional data into a higher dimensional feature space in which we define 
a distance measure from the spectral decomposition of the operator-valued kernel, suitable for 
functions and which allows the learning module to take into account the sequential nature 
of the data and the dependencies along the time-axis. 
Moreover, compared to RLSC, $\kappa$ the number of eigenfunctions in FRLSC can be seen as one more degree 
of freedom which can be used to improve performance when input data are complex and not represented by a 
vector in $\mathbb{R}^d$ as usual. 
Note that the usual scalar case (output space is $\mathbb{R}$ and then $\kappa = 1$) can be recovered from 
the functional case; for example when the operator-valued kernel is 
constructed from the identity operator or/and the output space is the space of constant functions.

\begin{table}[t]
\begin{center}
    
    \begin{tabular}{ccccccccc}
        \hline
          & C1& C2 & C3 & C4 & C5 & C6 & C7 & C8 \\
        \hline\hline

        C1 & 92  & 4     & 4.76  & 0 			& 5.27 	& 11.3 	& 6.89 		& 0  \\
        C2 & 0   & 52    & 0  	 & 14     & 0		 	& 2.7 	& 0 		  & 0  \\
        C3 & 0   & 20    & 76.2  & 0 		& 0    	& 0 		& 17.24		& 5 \\
        C4 & 0   & 16    & 0     & 66     & 0    	& 0 		& 0 		  & 0  \\
        C5 & 4   & 8     & 0  	 & 4  		& 84.21 & 0 		& 6.8		  & 0  \\
        C6 & 4   & 0     & 0     & 0 			& 10.52 & 86  	& 0 		  & 0 \\
        C7 & 0   & 0     & 0     & 8 			& 0    	& 0 		& 69.07 	& 0  \\
        C8 & 0   & 0     & 19.04 & 8 			& 0    	& 0 		& 0 		  & 95  \\
        \hline\hline
    \multicolumn{9}{c} {\emph{Total Recognition Rate = 77.56\%}} \\

    \hline
    \end{tabular}
\end{center}
\caption{Confusion Matrix obtained when using the Regularized Least Squares Classification (RLSC) algorithm.\vspace{0.25cm}}
    \label{tab2}
\end{table}
\begin{table}[t]
\begin{center}
    \begin{tabular}{ccccccccc}
        \hline
          & C1& C2 & C3 & C4 & C5 & C6 & C7 & C8  \\
        \hline\hline

        C1 & \textbf{100} & 0   & 0  	& 2    & 0 		& 5.3 	& 3.4 & 0  \\
        C2 & 0   & \textbf{82}  & 0  	& 8    & 0 		& 0 		& 0 	& 0  \\
        C3 & 0   & 14  & \textbf{90.9} & 8    & 0    & 0 		& 3.4 & 0 \\
        C4 & 0   & 4   & 0  	& \textbf{78}   & 0    & 0 		& 0 	& 0  \\
        C5 & 0   & 0   & 0  	& 1 	 & \textbf{89.47} & 0 		& 6.8 & 0  \\
        C6 & 0   & 0   & 0  	& 0 	 & 10.53 & \textbf{94.7}  & 0 	& 0 \\
        C7 & 0   & 0   & 0  	& 0 	 & 0 		& 0 		& \textbf{86.4} & 0  \\
        C8 & 0   & 0   & 9.1 	& 3	   & 0	  & 0 		& 0 	& \textbf{100}  \\
        \hline\hline
    \multicolumn{9}{c} {\emph{Total Recognition Rate = \textbf{90.18\%}}} \\

    \hline
    \end{tabular}
\end{center}
\caption{Confusion Matrix obtained when using the proposed Functional Regularized Least Squares Classification (FRLSC) algorithm.\vspace{0.35cm}}
    \label{tab1}
\end{table}

\section{Conclusion}

We have presented a learning methodology for nonlinear functional data analysis, which is an extension 
of scalar-valued and matrix-valued kernel based methodologies to the functional response setting. 
The problem of functional supervised learning is formalized as the problem of learning an operator 
between two infinite dimensional scalar-valued Hilbert spaces in a reproducing kernel Hilbert space of 
function-valued functions. 
We have introduced a set of rigorously defined operator-valued kernels that can be valuably applied 
to nonparametric operator learning when input and output data are continuous smooth functions, and we 
have showed their use for solving the problem of minimizing a regularized risk functional 
in the case of functional outputs without the need to discretize covariate and target functions. 
Our fully functional approach has been successfully applied to the problems of speech inversion and 
sound recognition, showing that the proposed framework is particularly relevant for audio signal processing 
applications where attributes are functions and dependent of each other. 

In future work, it would be interesting to explore further the potential of our proposed method in other 
machine learning problems such as  
collaborative filtering~\citep{Abernethy-2009} and structured output prediction~\citep{Brouard-2011, Kadri-2013} by building operator-valued kernels that can capture 
not only the functional information of responses, but also other types of output structure. 
In this context, learning the operator-valued kernel would be interesting to find the right model of dependencies 
between outputs. Recent works in this direction includes the papers of~\citet{Dinuzzo-2011}, \citet{Kadri-2012}, \citet{Sindhwani-2013} and \citet{Lim-2014}, but further investigations are needed in this area.
On the algorithmic side, possible extensions of this work include on-line implementations to deal with the 
case where the functional data set is made available step by step~\citep{Audiffren-2015}. Learning,~sequentially and 
without re-training from scratch at each iteration, a new function-valued function 
for each new observed pair of functional samples would be of practical interest.
Finally, although not covered in this paper, the analysis we present is likely to be applicable to learning problems that involve functional data with different functional profiles~(\eg, both smooth and spiky functions). Designing nonseparable operator-valued kernels that can exhibit better ability to characterize different smoothing levels is also an interesting future research direction.
\vspace{0.3cm}

\acks{%
We thank the anonymous reviewers for several
insightful comments that helped improve the original version of this paper.
We also thank M. Mbekhta for fruitful
discussions on the spectral theory of block operator matrices and A. Rabaoui 
for providing the sound recognition data set. 
A large part of this research was done while H.K. was at SequeL (INRIA-Lille) and J.A. at Aix-Marseille Universit\'e and LIF.
This work was supported by Ministry of Higher Education
and Research, Nord-Pas-de-Calais Regional
Council and FEDER through the `Contrat de Projets
Etat Region (CPER) 2007-2013'. 
H.K. acknowledges the support of a 
Junior Researcher Contract No. 4297 from
the Nord-Pas-de-Calais region.
H.K. and P.P. acknowledge the support of the French 
National Research Agency (ANR-09-EMER-007 project LAMPADA). 
H.K. was also supported in part by French grants from the CNRS-LAGIS~(ANR KernSig Project). 
A.R. was supported by
the PASCAL2 Network of Excellence, ICT-216886,
ANR Project ASAP ANR-09-EMER-001 and the INRIA
ARC MABI.
P.P. also acknowledges the support of INRIA.
}


\appendix

\section{Proof of Theorem~\ref{th:positiveDefiniteImpliesKernel} - Completion of  $\mathcal{F}_0$}

\label{App:completion}

We show below how to construct the Hilbert space
$\mathcal{F}$ of $\mathcal{Y}$-valued functions, that is the
completion of the function-valued pre-Hilbert space $\mathcal{F}_{0}$.
$\mathcal{F}_{0}\subset\mathcal{Y}^\mathcal{X}$ is the space of all
$\mathcal{Y}$-valued functions $F$ of the form
$F(\cdot)=\sum_{i=1}^{n}K(w_{i},\cdot)u_{i}$, where
$w_{i} \in \mathcal{X}$ and $u_{i} \in \mathcal{Y}$,
$i=1,\ldots, n$.  Consider the inner product of the
functions $F(\cdot)=\sum_{i=1}^{n}K(w_{i},\cdot)u_{i}$ and 
$G(\cdot)=\sum_{j=1}^{m}K(z_{j},\cdot)v_{j}$ from $\mathcal{F}_{0}$ defined as follows
\begin{align*}
\langle F(\cdot),G(\cdot)\rangle _{\mathcal{F}_{0}}=\langle \sum\limits_{i=1}^{n}K(w_{i},\cdot)u_{i},\sum\limits_{j=1}^{m}K(z_{j},\cdot) 
v_{j}\rangle _{\mathcal{F}_{0}} 
 = \sum\limits_{i=1}^{n} \sum\limits_{j=1}^{m} \langle 
K(w_{i},z_{j})u_{i},v_{j}\rangle _{\mathcal{Y}}.
\end{align*}
We have shown that $(\mathcal{F}_{0},\langle .,.\rangle _{\mathcal{F}_{0}})$ is
a pre-Hilbert space. This pre-Hilbert space is in general not
complete, but It can be completed via Cauchy sequences to build the $\mathcal{Y}$-valued reproducing kernel Hilbert space $\mathcal{F}$.

Consider any Cauchy sequence $\{F_{n}(\cdot)\} \subset \mathcal{F}_{0}$,
for every $w \in \mathcal{X}$ the functional $F(w)$ {is bounded}, since
\begin{eqnarray*}
  \begin{array}{ll}
    \|F(w)\|_{\mathcal{Y}} &= \sup\limits_{\|u\|=1} | \langle F(w), u \rangle _{\mathcal{Y}}|
    =   \sup\limits_{\|u\|=1} | \langle F(\cdot),K(w,\cdot) u \rangle _{\mathcal{F}_{0}} | \\[0.5cm]
    & \leq \|F(\cdot)\|_{\mathcal{F}_{0}} \sup\limits_{\|u\|=1} \|K(w,\cdot) u \|_{\mathcal{F}_{0}}
    \leq \|F(\cdot)\|_{\mathcal{F}_{0}} \sup\limits_{\|u\|=1} \sqrt{\langle K(w,w) u, u \rangle_{\mathcal{Y}}} \\[0.5cm]
     & \leq M_{w}\|F(\cdot)\|_{\mathcal{F}_{0}} \mbox{ with }
    M_{w}= \sup\limits_{\|u\|=1} \sqrt{\langle K(w,w)u,u\rangle _{\mathcal{Y}}}.
  \end{array}
\end{eqnarray*}
Moreover, if the kernel $K$ is Mercer, it is locally bounded~(see~\citealp[Proposition~2]{Carmeli-2010}). It is easy to see that in this case $  \|F(w)\|_{\mathcal{Y}}  \leq M \|F(\cdot)\|_{\mathcal{F}_{0}}$, where $M$ here does not depend on $w$.
Consequently, 
\begin{equation*}\|F_{n}(w)-F_{m}(w)\|_{\mathcal{Y}}\leq
M \|F_{n}(\cdot)-F_{m}(\cdot)\|_{\mathcal{F}_{0}}.
\end{equation*}
It follows that $\{F_{n}(w)\}$ is a Cauchy sequence in
$\mathcal{Y}$ and by the completeness of the space
$\mathcal{Y}$, there exists a $\mathcal{Y}$-valued function
$F$ where, $\forall w\in \mathcal{X}$,
$F(w)=\lim\limits_{n\rightarrow \infty} F_{n}(w)$. So the Cauchy
sequence $\{F_{n}(\cdot)\}$ defines a function $F(\cdot)$ to which it is
convergent at every point of $\mathcal{X}$.

Let us denote $\mathcal{F}$ the linear space containing all the
functions $F(\cdot)$, the limits of Cauchy sequences $\{F_{n}(\cdot)\} \subset
\mathcal{F}_{0}$, and consider the 
norm in $\mathcal{F}$ defined by 
  $\|F(\cdot)\|_{\mathcal{F}}=\lim\limits_{n\rightarrow
    \infty}\|F_{n}(\cdot)\|_{\mathcal{F}_{0}}$, 
where ${F_{n}(\cdot)}$ is a Cauchy sequence of $\mathcal{F}_{0}$
converging to $F(\cdot)$. This norm is well defined since it does not
depend on the choice of the Cauchy sequence. In fact, suppose that two
Cauchy sequences $\{F_{n}(\cdot)\}$ and $\{G_{n}(\cdot)\}$ in
$\mathcal{F}_{0}$ define the same function $F(\cdot)\in \mathcal{F}$.
Then $\{F_{n}(\cdot)-G_{n}(\cdot)\}$ is also a Cauchy sequence and $\forall
w\in \mathcal{X},\ \lim\limits_{n\rightarrow
  \infty}F_{n}(w)-G_{n}(w)=0$. Hence, $\lim\limits_{n\rightarrow
  \infty}\langle F_{n}(w)-G_{n}(w), u \rangle _{\mathcal{Y}}=0$ for any
$u \in \mathcal{G}_{y}$ and using the reproducing property, it
follows that $\lim\limits_{n\rightarrow
  \infty}\langle F_{n}(\cdot)-G_{n}(\cdot),H(\cdot)\rangle _{\mathcal{F}_{0}}=0$ for any
function $H(\cdot)\in \mathcal{F}_{0}$ and thus 
\begin{align*}
\lim\limits_{n\rightarrow
  \infty}\|F_{n}(\cdot)-G_{n}(\cdot)\|_{\mathcal{F}_{0}}=0.
\end{align*}
Consequently,
\begin{eqnarray*}
  |\lim\limits_{n\rightarrow
    \infty}\|F_{n}\|-\lim\limits_{n\rightarrow \infty}\|G_{n}\|\
  | =\lim\limits_{n\rightarrow \infty}|\ \|F_{n}\|-\|G_{n}\|\ | \leq
  \lim\limits_{n\rightarrow \infty}\|F_{n}-G_{n}\|=0.
\end{eqnarray*}
So that for any function $F(\cdot)\in \mathcal{F}$ defined by two
different Cauchy sequences $\{F_{n}(\cdot)\}$ and $\{G_{n}(\cdot)\}$ in
$\mathcal{F}_{0}$, we have $\lim\limits_{n\rightarrow
  \infty}\|F_{n}(\cdot)\|_{\mathcal{F}_{0}}=\lim\limits_{n\rightarrow
  \infty}\|G_{n}(\cdot)\|_{\mathcal{F}_{0}}=\|F(\cdot)\|_{\mathcal{F}}$.
$\|.\|_{\mathcal{F}}$ has all the properties of a norm and defines in
$\mathcal{F}$ an inner product which on $\mathcal{F}_{0}$ coincides
with $\langle .,.\rangle _{\mathcal{F}_{0}}$ already defined. It remains to be shown
that $\mathcal{F}_{0}$ is dense in $\mathcal{F}$ which is a complete
space.

For any $F(\cdot)$ in $\mathcal{F}$ defined by the Cauchy sequence
$F_{n}(\cdot)$, we have
  $\lim\limits_{n\rightarrow
    \infty}\|F(\cdot)-F_{n}(\cdot)\|_{\mathcal{F}}=\lim\limits_{n\rightarrow
    \infty}\lim\limits_{m\rightarrow
    \infty}\|F_{m}(\cdot)-F_{n}(\cdot)\|_{\mathcal{F}_{0}} = 0$. 
It follows that $F(\cdot)$ is a strong limit of $F_{n}(\cdot)$ in
$\mathcal{F}$ and then $\mathcal{F}_{0}$ is dense in $\mathcal{F}$. 
To prove that $\mathcal{F}$ is a complete space, we consider
$\{F_{n}(\cdot)\}$ any Cauchy sequence in $\mathcal{F}$. Since
$\mathcal{F}_{0}$ is dense in $\mathcal{F}$, there exists a sequence
$\{G_{n}(\cdot)\}\subset \mathcal{F}_{0}$ such that
$\lim\limits_{n\rightarrow
  \infty}\|G_{n}(\cdot)-F_{n}(\cdot)\|_{\mathcal{F}}=0$. Besides
$\{G_{n}(\cdot)\}$ is a Cauchy sequence in $\mathcal{F}_{0}$ and thus
defines a function $H(\cdot)\in \mathcal{F}$ which verifies
$\lim\limits_{n\rightarrow \infty}\|G_{n}(\cdot)-h(\cdot)\|_{\mathcal{F}}=0$.
So $\{G_{n}(\cdot)\}$ converges strongly to $H(\cdot)$ and then $\{F_{n}(\cdot)\}$
also converges strongly to $H(\cdot)$ which means that the space
$\mathcal{F}$ is complete. In addition, $K(\cdot,\cdot)$ has the
reproducing property in $\mathcal{F}$. To see this, let $F(\cdot)\in
\mathcal{F}$, then $F(\cdot)$ is defined by a Cauchy sequence $\{F_{n}(\cdot)\}
\subset \mathcal{F}_{0}$ and we have {from the continuity of the inner product in $\mathcal{F} \subset\mathcal{Y^X}$ (endowed with the uniform topology) that, for all $w\in \mathcal{X}$
and $u \in \mathcal{Y}$,
\begin{eqnarray*}
  \begin{array}{lll}
    \langle F(w), u \rangle _{\mathcal{Y}}&=&\langle \lim\limits_{n\rightarrow
      \infty}F_{n}(w), u \rangle _{\mathcal{Y}}
     = \lim\limits_{n\rightarrow
      \infty}\langle F_{n}(w), u \rangle _{\mathcal{Y}} 
    =\lim\limits_{n\rightarrow \infty}
    \langle F_{n}(\cdot),K(w,\cdot) u \rangle_{\mathcal{F}_{0}}\\
    &=& \langle \lim\limits_{n\rightarrow
      \infty}F_{n}(\cdot),K(w,\cdot) u \rangle _{\mathcal{F}}
    =\langle F(\cdot),K(w,\cdot) u \rangle _{\mathcal{F}}.
  \end{array}
\end{eqnarray*}
Finally, we conclude that $\mathcal{F}$ is a reproducing kernel
Hilbert space since $\mathcal{F}$ is a real inner product space that
is complete under the norm $\|.\|_{\mathcal{F}}$ defined above, and
has $K(\cdot,\cdot)$ as reproducing kernel.
\hfill $\blacksquare$


\section{Representer Theorem}
\label{App:RepresenterTheorem}

We provide here a proof of the analog of the representer theorem in the case of function-valued reproducing kernel Hilbert spaces.
\begin{theo}
  \label{th:representerTh} (representer theorem) \vspace{1mm} \\
  Let $K$ a nonnegative Mercer operator-valued kernel and
  $\mathcal{F}$ its corresponding function-valued reproducing kernel
  Hilbert space.  The solution $\widetilde{F}_\lambda\in\mathcal{F}$ of the
  regularized optimization problem
  \begin{equation*}
    \widetilde{F}_\lambda = \arg\min_{F \in \mathcal{F}}\sum\limits_{i=1}^{n}\|y_{i}-F(x_{i})\|_{\mathcal{Y}}^{2}+\lambda\|F\|_{\mathcal{F}}^{2}
  \end{equation*}
  has the following form
  \begin{eqnarray}
    \nonumber
    \widetilde{F}_\lambda(\cdot)=\sum\limits_{i=1}^{n}K(x_{i},\cdot)u_{i},
  \end{eqnarray}
  where $u_i \in \mathcal{Y}$.
\end{theo}

\textbf{Proof}: 
We use the Frechet derivative which is the strongest notion of derivative in a normed linear space; see, for example, Chapter~4 of~\citet{Kurdila-2005}. We use the standard notation $D_F$ for the Frechet derivative operator.
Let $J_{\lambda}(F) = \displaystyle\sum\limits_{i=1}^{n}\|y_{i}-F(x_{i})\|_{\mathcal{Y}}^{2}+\lambda\|F\|_{\mathcal{F}}^{2}$ 
be the functional to be minimized. 
$\widetilde{F}$ is the operator in $\mathcal{F}$ such that 
$\widetilde{F} = \arg\min\limits_{F \in \mathcal{F}} J_{\lambda}(F) \Rightarrow D_F J_{\lambda}(\widetilde{F})=0$. 
To compute $D_F J_{\lambda}(F)$, we use the Gateaux derivative $D_G$ of $J_{\lambda}$ with respect to $F$ in the direction $H$, which is defined by:
\begin{equation}
  \nonumber D_{G}J_{\lambda}(F,H) = \lim\limits_{\tau \longrightarrow
    0} \frac{J_{\lambda}(F+\tau H) - J_{\lambda}(F)}{\tau}.
\end{equation}
$J_{\lambda}$ can be written as
$J_{\lambda}(F)=\displaystyle\sum\limits_{i=1}^{n}G_{i}(F)+\lambda L(F)$
and using the fact that  $D_{G}J_{\lambda}(F,H) = \langle D_{G}J_{\lambda}(F),H\rangle $ we obtain 
\begin{enumerate}[i.]
  \item \label{enum3:i} $L(F)=\|F\|_{\mathcal{F}}^{2}$

    $\lim\limits_{\tau \longrightarrow 0} \displaystyle \frac{\|F+\tau
      H\|_{\mathcal{F}}^{2}-\|F\|_{\mathcal{F}}^{2}}{\tau} = 2\langle F,H\rangle \ $
    $\Longrightarrow \   D_{G}L(F)=2F$.

  \item \label{enum3:ii} $G_{i}(F)=\|y_{i}-F(x_{i})\|_{\mathcal{Y}}^{2}$

 $ \displaystyle \lim\limits_{\tau \longrightarrow
        0} \displaystyle \frac{\|y_{i}-F(x_{i})-\tau
        H(x_{i})\|_{\mathcal{Y}}^{2}-\|y_{i}-F(x_{i})\|_{\mathcal{Y}}^{2}}{\tau} 
       = -2\langle y_{i}-F(x_{i}),H(x_{i})\rangle _{\mathcal{Y}}$ 
\vspace{2mm}
\\
      $ = -2\langle K(x_{i},\cdot)(y_{i}-F(x_{i})),H\rangle _{\mathcal{F}}
       = -2\langle K(x_{i},\cdot)u_{i},H\rangle _{\mathcal{F}} $
       with $u_{i}=y_{i}-F(x_{i})$

    $\Longrightarrow D_{G}G_{i}(F)=-2K(x_{i},\cdot)u_{i}$.
\end{enumerate}
When the kernel $K$ is Mercer, Corollary~4.1.1 in~\citet{Kurdila-2005} can be applied to show that $J_{\lambda}$ is Fr\'echet differentiable and that, $\forall F \in \mathcal{F}$, $D_{F}J_{\lambda}(F) = D_{G}J_{\lambda}(F)$.
Using (\ref{enum3:i}), (\ref{enum3:ii}), and $D_{F}J_{\lambda}(\widetilde{F})=0
\ \Longrightarrow \ \widetilde{F}(\cdot)=\displaystyle
\frac{1}{\lambda}\sum\limits_{i=1}^{n}K(x_{i},\cdot)u_{i}$.
\hfill $\blacksquare$


\section{Proof of Theorem~\ref{beta stable}}\label{app:stability}

We show here that under Assumptions \ref{Hyp K bounded}, \ref{Measure} and \ref{Hyp on c}, a learning algorithm that maps a training set $Z$ to the function $F_Z$ defined in~(\ref{definition fZ}) is $\beta$ stable with $\beta= \displaystyle\frac{\sigma^2 \kappa^2 }{2n\lambda }$.
First, since $\ell$ is convex with respect to $F$, we have $\forall 0 \le t \le 1$
\begin{equation}\nonumber
\begin{aligned}
\ell(y,F_Z + t(F_\zi-F_Z),x) -\ell(y,F_Z,x)  \le t \left(\ell(y,F_\zi,x) - \ell(y,F_Z,x) \right).
\end{aligned}
\end{equation}
Then, by summing over all couples $(x_k,y_k)$ in $\zi$,
\begin{equation}\label{eq Rempun}
\begin{aligned}
R_{emp}(F_Z + t(F_\zi-F_Z),\zi) -R_{emp}(F_Z ,\zi)  \le t \left(R_{emp}(F_\zi,\zi) -R_{emp}(F_Z ,\zi) \right).
\end{aligned}
\end{equation}
Symmetrically, we also have
\begin{equation}\label{eq Rempdeux}
\begin{aligned}
R_{emp}(F_\zi + t(F_Z-F_\zi),\zi) -R_{emp}(F_\zi ,\zi)  \le t \left(R_{emp}(F_Z,\zi) -R_{emp}(F_\zi ,\zi) \right).
\end{aligned}
\end{equation}
Thus, by summing \eqref{eq Rempun} and \eqref{eq Rempdeux}, we obtain
\begin{equation}\label{eq Remptrois}
\begin{aligned}
R_{emp}(F_Z & + t(F_\zi-F_Z),\zi) -R_{emp}(F_Z ,\zi)\\
& + R_{emp}(F_\zi + t(F_Z-F_\zi),\zi) -R_{emp}(F_\zi ,\zi)  \le 0.
\end{aligned}
\end{equation}
Now, by definition of $F_Z$ and $F_\zi$,
\begin{equation}\label{eq fmin}
\begin{aligned}
R_{reg}(F_Z ,Z) & - R_{reg}(F_Z + t(F_\zi-F_Z),Z)\\
& +R_{reg}(F_\zi ,\zi)-  R_{reg}(F_\zi + t(F_Z-F_\zi),\zi)  \le 0.
\end{aligned}
\end{equation}
Combining \eqref{eq Remptrois} and \eqref{eq fmin}, we find
\begin{equation}
\label{eq:ineq}
\begin{aligned}
&\ell(y_i,F_Z,x_i) - \ell(y_i,F_Z + t(F_\zi-F_Z),x_i)\\
&\hspace{0.5cm} + n \lambda \left( \| F_Z \|^2_\mathcal{F} - \| F_Z + t(F_\zi-F_Z)\|^2_\mathcal{F}
+ \| F_\zi \|^2_\mathcal{F} -\| F_\zi + t(F_Z-F_\zi) \|^2_\mathcal{F} \right)\le 0.
\end{aligned}
\end{equation}
Moreover, we have
\begin{align}
\label{eq:eq}
\| F_Z \|^2_\mathcal{F} & - \| F_Z + t(F_\zi-F_Z)\|^2_\mathcal{F} + \| F_\zi \|^2_\mathcal{F} -\| F_\zi + t(F_Z-F_\zi) \|^2_\mathcal{F} \nonumber \\
& =  \| F_Z \|^2_\mathcal{F} - \| F_Z \|^2_\mathcal{F} - t^2  \| F_\zi-F_Z\|^2_\mathcal{F} -2t \langle F_Z, F_\zi - F_z \rangle_\mathcal{F} \nonumber \\
& +  \| F_\zi \|^2_\mathcal{F} - \| F_\zi \|^2_\mathcal{F} - t^2  \| F_Z-F_\zi\|^2_\mathcal{F} -2t \langle F_\zi, F_Z - F_\zi \rangle_\mathcal{F} \nonumber \\
& = - 2 t^2  \| F_\zi-F_Z\|^2_\mathcal{F} -2t \langle F_Z, F_\zi - F_z \rangle_\mathcal{F}  -2t \langle F_\zi, F_Z - F_\zi \rangle_\mathcal{F} \nonumber \\
&  = - 2 t^2  \| F_\zi-F_Z\|^2_\mathcal{F} + 2 t  \| F_\zi-F_Z\|^2_\mathcal{F} \nonumber \\
& = 2t(1-t) \| F_\zi-F_Z\|^2_\mathcal{F}.
\end{align}
Hence, since $\ell$ is $\sigma$-Lipschitz continuous with respect to $F(x)$, we obtain from (\ref{eq:ineq}) and (\ref{eq:eq}), $\forall t \in \mathopen{]}0,1 \mathclose{[} $,
\begin{equation}\nonumber
\begin{aligned}
 \|F_Z-F_\zi \|^2_\mathcal{F} & \le  \frac{1}{2t(1-t)}  \left( \| F_Z \|^2_\mathcal{F} - \| F_Z + t(F_\zi-F_Z)\|^2_\mathcal{F} + \| F_\zi \|^2_\mathcal{F} -\| F_\zi + t(F_Z-F_\zi) \|^2_\mathcal{F} \right)\\
&\le \frac{1}{2t(1-t)n\lambda} \big( \ell(y_i,F_Z + t(F_\zi-F_Z),x_i)-\ell(y_i,F_Z,x_i)\big) \\
&\le \frac{\sigma }{2(1-t)n\lambda} \|F_\zi(x_i) - F_Z(x_i)\|_\y.
\end{aligned}
\end{equation}
In particular, when $t$ tends to $0$, we have
$$ \|F_Z-F_\zi \|^2_\mathcal{F} \le \frac{\sigma }{2n\lambda} \|F_\zi(x_i) - F_Z(x_i)\|_\y \le \frac{\sigma \kappa }{2n\lambda} \|F_\zi- F_Z\|_\mathcal{F},$$
which gives that $$\displaystyle \|F_Z-F_\zi \|_\mathcal{F} \le  \frac{\sigma \kappa }{2n\lambda}.$$
This implies that, $\forall (x,y)$,
\begin{equation}\nonumber
\vert \ell(y,F_Z,x) - \ell(y,F_\zi,x) \vert \le \sigma \|F_Z(x) - F_\zi(x)\|_\y \le \sigma \kappa \|F_Z-F_\zi \|_\mathcal{F} \le \frac{\sigma^2 \kappa^2 }{2n\lambda},
\end{equation}
which concludes the proof.
\hfill $\blacksquare$

\section{Proof of Lemma~\ref{Hyp LSR}}\label{app:admissible}

We show here that Assumption~\ref{c control} is satisfied for the least squares loss function when Assumption~\ref{Hyp Y bounded} holds and use that to prove Lemma~\ref{Hyp LSR}}.
First, note that $\ell$ is convex with respect to its second argument. Since $\mathcal{F}$ is a vector space, $0 \in \h$. Thus,  
\begin{equation}\label{astuce 0}
  \lambda \|F_Z\|^2 \le R_{reg}(F_Z,Z) \le  R_{reg}(0,Z) \le \frac{1}{n}\sum_{k=1}^n \|y_k \|^2 \le \sigma_y ^2, 
\end{equation}
where we used the definition of $F_Z$ (see Equation~\ref{definition fZ}) and the bound on $Y$ (Assumption~\ref{Hyp Y bounded}). This inequality is uniform over $Z$, and thus holds for $F_\zi$.
Moreover,
$\forall x \in \x$, 
\begin{equation}\nonumber
\|F_Z(x)\|_\y^2 = \left\langle F_Z(x), F_Z(x)\right\rangle_\y =\left\langle K(x,x) F_Z, F_Z\right\rangle_\mathcal{F} \le \|K(x,x)\|_{op} \|F_Z\|^2_\mathcal{F} \le \kappa^2 \frac{\sigma_y^ 2}{\lambda} .
\end{equation}
Hence, using Lemma \ref{Lem reprod} and (\ref{astuce 0}), we obtain 
\begin{equation*}\label{RR hyp4}
 \|y - F_Z(x)\|_\y \le  \|y \|_\y +  \|f_Z(x)\|_\y \le \sigma_y + \kappa \frac{\sigma_y}{\sqrt{\lambda}}, \ \forall (x,y)\in \x \times \y.
\end{equation*}
Then it follows that
\begin{equation}\nonumber
\begin{aligned}
\big\vert &\|y - F_Z(x)\|_\y^2 -  \|y - F_\zi(x)\|_\y^2 \big\vert\\
&= \big\vert \|y - F_Z(x)\|_\y -  \|y - F_\zi(x)\|_\y \big\vert \ \big\vert \|y - F_Z(x)\|_\y +  \|y - F_\zi(x)\|_\y  \big\vert\\
&\le 2 \sigma_y ( 1 + \frac{\kappa}{\sqrt{\lambda}}) \|F_Z(x)-F_\zi(x)\|_\y .
\end{aligned}
\end{equation}
\hfill $\blacksquare$

\section{Proof of Remark~\ref{remarkHS}}\label{app:HS}

We show here that if the finite trace assumption of the operator $K(x,x)$ in~\citet{Caponnetto-2006} is satisfied,  then our Assumption~\ref{Hyp K bounded} on the kernel holds.
Let $K$ be an operator-valued kernel satisfying the hypotheses of \citet{Caponnetto-2006}, i.e $K_x$ is Hilbert-Schmidt and $\sup_{x \in \x} Tr(K(x,x)) < + \infty $. 
Then, $\exists \eta >0$, $\forall x\in \x$, $\exists \left(e_j^x\right)_{j\in\N}$ an orthonormal basis of $\y$, $\exists \left(h_j^x\right)_{j\in\N}$ an orthogonal family of $\mathcal{F}$ with $\sum_{j\in\N} \|h_j^x\|_\mathcal{F}^2 \le \eta$ such that $\forall y \in \y,$
$$K(x,x)y=\sum_{j,\l} \left\langle h_j^x,h_\l^x \right\rangle_\mathcal{F}\left\langle y,e_j^x \right\rangle_\y e_\l^x. $$
Thus, $\forall i \in \N$,
\begin{equation}\nonumber
\begin{aligned}
K(x,x)e_i^x =\sum_{j,\l} \left\langle h_j^x,h_\l^x \right\rangle_\mathcal{F}\left\langle e_i^x,e_j^x \right\rangle_\y e_\l^x 
 =\sum_\l \left\langle h_i^x,h_\l^x \right\rangle_\mathcal{F} e_\l^x.
\end{aligned}
\end{equation}
Hence 
\begin{equation}\nonumber
\begin{aligned}
\|K(x,x)\|^2_{op} &=\sup_{i\in\N} \|K(x,x)e_i^x\|_\y^2 
=\sup_{i\in\N}\sum_{j,\l} \left\langle h_i^x,h_\l^x \right\rangle_\mathcal{F} \left\langle h_i^x,h_j^x \right\rangle_\mathcal{F} \left\langle e_j^x,e_\l^x \right\rangle_\y\\
&=\sup_{i\in\N}\sum_{\l} (\left\langle h_i^x,h_\l^x \right\rangle_\mathcal{F})^2 \le \sup_{i\in\N}\|h_i^x\|_\mathcal{F}^2 \sum_{\l} \|h_\l^x\|_\mathcal{F}^2 \le \eta^2.
\end{aligned}
\end{equation}
 \hfill $\blacksquare$


\bibliography{HK_JMLR_v3}
\end{document}